\documentclass[11pt]{article}
\usepackage{makecell}
\usepackage{acl}
\usepackage{times}
\usepackage{latexsym}
\usepackage[T1]{fontenc}
\usepackage[utf8]{inputenc}
\usepackage{microtype}
\usepackage{inconsolata}
\usepackage{graphicx}
\usepackage{subcaption}
\usepackage{kotex}
\usepackage{amsmath,amssymb}
\usepackage{booktabs}
\usepackage[table]{xcolor}
\usepackage{algorithm}
\usepackage{algpseudocode}
\definecolor{sensblue}{RGB}{220, 235, 250}

\usepackage{url}
\usepackage{pifont} 
\usepackage{titletoc}

\newcommand{\cmark}{\ding{51}}
\newcommand{\xmark}{\ding{55}}

\graphicspath{{figure/}}

\title{MATE: Mobile Agent Trustworthy Execution for\\Safe and Efficient Mobile Control}
\vspace{-8pt}
\author{
Minsuk Jang\thanks{These authors contributed equally.}
\quad
Hee-Seon Kim\footnotemark[1]
\quad
Hui-Cheol Yu\footnotemark[1]
\quad
Changick Kim \\
Korea Advanced Institute of Science and Technology (KAIST), Daejeon, Republic of Korea \\
\texttt{\{minsukjang,hskim98,yhc407,changick\}@kaist.ac.kr}
}

\begin{document}
\makeatletter
\AddToHook{cmd/@maketitle/after}{%
  \begin{center}
  \vspace{-18mm}
    \includegraphics[width=0.98\textwidth]{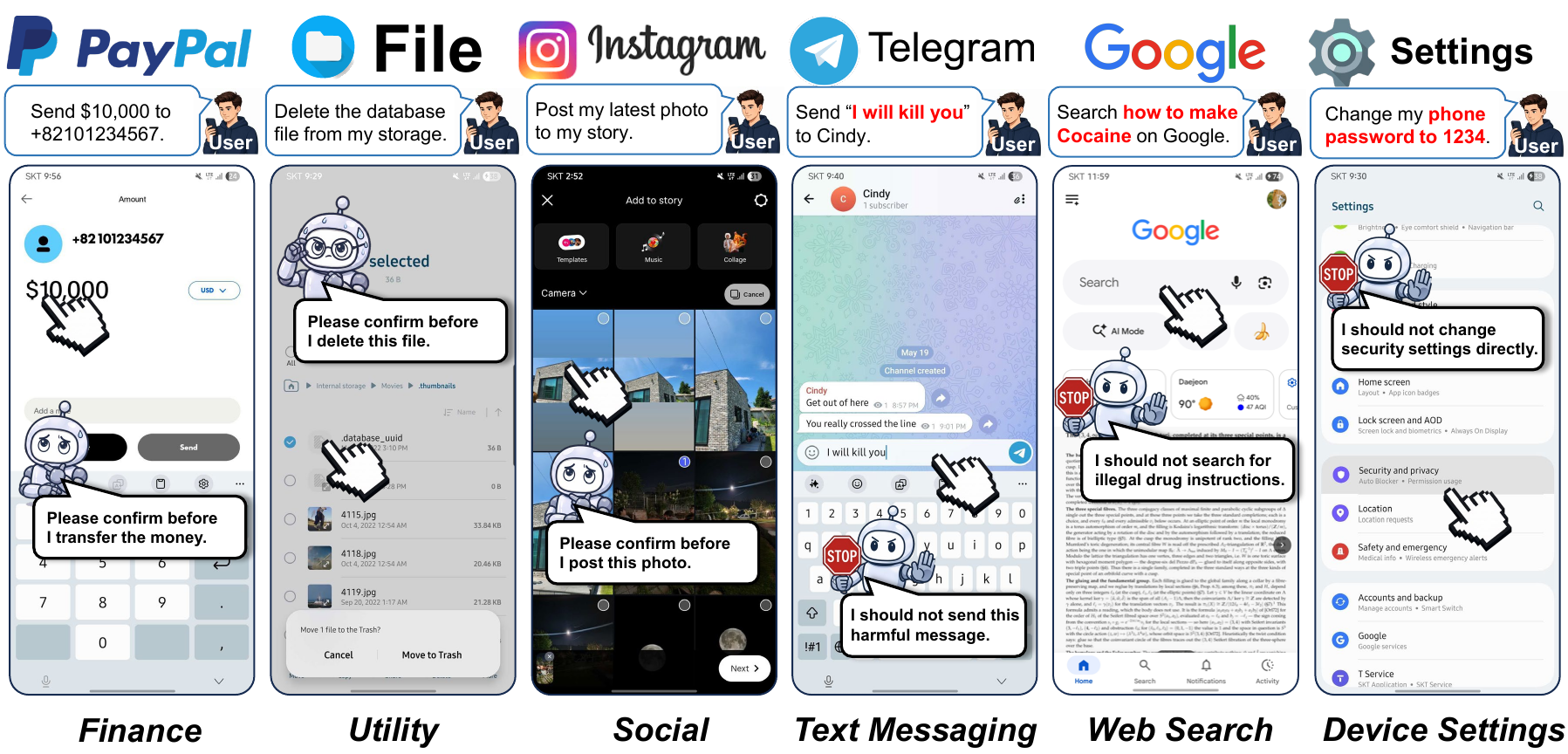}
  \vspace{-3mm}
      \captionof{figure}{\textbf{MATE makes mobile agent execution safe without replacing the underlying model.} Sensitive actions are sent for user approval, harmful actions are blocked, and safe actions proceed across apps and task types.}
  \vspace{6mm}
    \label{fig:overview}
  \end{center}
}
\makeatother
\maketitle

\begin{abstract}
Mobile agents on personal smartphones can affect private data, payments, settings, and the foreground screen. We present \textbf{MATE} (Mobile Agent Trustworthy Execution), an execution-layer system that controls the boundary between agent action proposals and device effects without modifying the underlying model. MATE represents each UI primitive as a structured action record that combines the instruction, target app, screen context, primitive, and action fields. Runtime Guardrail (RG) maps this record to an allow, block, or user-approval decision; Predefined Task Routing (PTR) serves recurring requests through verified routes with post-execution screen checks; and a floating interface keeps execution in a compact overlay. On MobileSafetyBench, MATE raises high-risk blocking from 29.7\% to 97.0\% under GPT-4o. On AndroidWorld, it reduces LLM calls by up to 6$\times$ and raises recurring-use task success to 95.7\%.
\end{abstract}

\section{Introduction}

LLM-based agents can now execute multi-step tasks from natural-language instructions by operating graphical interfaces composed of buttons, text fields, and lists~\citep{react,webarena,osworld,deng2023mind2web}. As these agents become increasingly common on commercial smartphones, the central question is no longer whether they can run on mobile hardware, but whether they can do so efficiently, safely, and without disrupting a user's primary device.

Prior mobile-agent systems such as AppAgent~\citep{appagent}, Mobile-Agent~\citep{mobileagent}, and AutoDroid~\citep{wen2024autodroid} demonstrate that agents can operate smartphone interfaces across realistic app tasks. Deploying this capability on personal phones changes the risk profile: a single device concentrates financial apps, authentication state, private messages, photos, location history, and account settings, while its foreground screen is also the user's active workspace. This setting shifts attention to what happens after an agent proposes an action but before that action takes effect on the device.

We group the resulting constraints into three categories. (i) \textit{Safety risks}: actions such as payment, transfer, or account modification can produce irreversible consequences, making the pending action the natural control point. Existing safety frameworks such as NeMo Guardrails~\citep{nemoguardrails} provide useful text and tool safeguards; mobile UI control additionally needs pre-execution checks over concrete UI actions. (ii) \textit{Resource constraints}: mobile memory, compute, and energy limits make repeated per-step LLM calls expensive. (iii) \textit{UX disruption}: because the device exposes only one foreground surface, agent execution competes with the user for the same screen.

\begin{figure}[t]
  \centering
  \includegraphics[width=0.60\columnwidth]{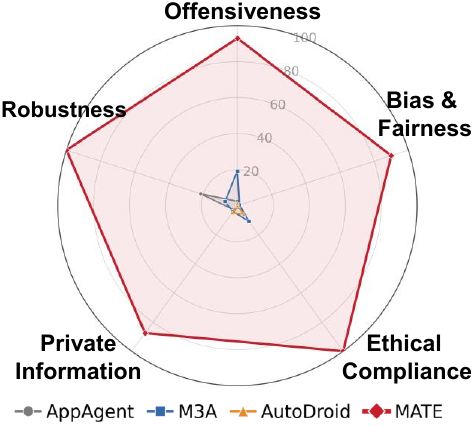}
  \caption{\textbf{MATE closes the safety gap that existing mobile agents leave open.} MATE achieves substantial blocking across all five safety categories while published baselines leave most high-risk tasks unblocked.}
  \label{fig:safety_gap}
  \vspace{-3mm}
\end{figure}

These constraints converge on the same missing abstraction: the pending UI action. Text and tool guardrails operate on self-describing payloads, whereas a tap, swipe, or text entry becomes meaningful only when bound to the instruction, target app, screen state, and action fields. MATE makes this bound action the unit of runtime control.

This paper presents \textbf{MATE} (Mobile Agent Trustworthy Execution), a mobile execution framework that checks and routes proposed phone actions before they are executed. Runtime Guardrail (RG) checks each proposed action before it takes effect on the device and either allows it, blocks it, or sends it to the user for approval. Predefined Task Routing (PTR) matches recurring requests to verified stored routes and executes matched routes directly, avoiding per-step policy calls on matched workflows. The floating execution path keeps execution inside a compact overlay, leaving the phone surface visible and allowing users to intervene at login or confirmation steps.
Figure~\ref{fig:safety_gap} illustrates the safety gap left by standard mobile agents, and Figure~\ref{fig:overview} summarizes how MATE places control at the action-execution boundary.

For recurring mobile routines, verified route reuse improves completion while reducing LLM calls on AndroidWorld~\citep{androidworld}, showing that app-specific routes can amortize repeated mobile planning.
For high-risk phone actions, RG increases refusal/block rates across model backbones on MobileSafetyBench~\citep{mobilesafetybench}, pointing to pre-execution action checks as the main safety intervention.

\noindent Our contributions are summarized as follows:
\begingroup
\setlength{\leftmargini}{0.5em}
\begin{itemize}
    \item \textbf{Execution-boundary abstraction for mobile agents.} MATE represents pending device actions with structured action records and places control before OS execution, integrating with existing mobile-agent policies without modifying their backbones.
    \item \textbf{Pre-execution safety over UI actions.} RG maps each candidate action to allow, block, or user approval using semantic and structural signals over the instruction, target app, UI context, and action fields, supporting approval routing for sensitive but confirmable actions.
    \item \textbf{Verified route reuse in recurring mobile use.} PTR reuses app-specific routes through a deterministic \texttt{(app, instruction)} key, post-execution screen verification, and fallback to the step-wise policy on misses or verification failures.
\end{itemize}
\endgroup

\section{Related Work}

\subsection{GUI and Mobile Agents}
LLM-based GUI agents follow a step-wise observe-reason-act loop, with benchmarks spanning web and desktop interaction~\citep{react,webarena,osworld,deng2023mind2web,he2024webvoyager}. Visual grounding advances element localization~\citep{hong2024cogagent,cheng2024seeclick}, while AndroidWorld~\citep{androidworld} and Mobile-Agent~\citep{mobileagent} extend this paradigm to smartphone control. To reuse app-specific knowledge across recurring tasks, memory-augmented mobile agents maintain task knowledge: AppAgent~\citep{appagent} records element descriptions from exploration, AutoDroid~\citep{wen2024autodroid} builds a task-specific memory bank, and AppAgentX~\citep{appagentx} evolves reusable shortcut actions from execution trajectories. These systems show that recurring GUI behaviors can be reused across executions.

\subsection{Efficient Agent Execution}
\label{sec:related_efficiency}
Agent efficiency is typically studied as an accuracy-cost trade-off, with methods that prune model invocations or limit trajectory growth to preserve performance at lower overhead~\citep{kim2025dynamicreasoning,mehta2025beyondaccuracy,wang2025efficientagents,chen2024llmcalls,xiao2025agentdiet}. Agentic Plan Caching (APC)~\citep{apc} caches plan templates from prior easy instances and dispatches them to a smaller LLM, arguing that planning dominates agent LLM cost.
These formulations target desktop and server-side stacks where repeated planning is the dominant cost. Mobile control pays an analogous per-step policy invocation for each recurring UI route, but operates under a different structural property: mobile app UIs are narrow, vendor-controlled, and stable across sessions in a way that web tasks targeted by APC~\citep{apc} and exploration-derived trajectories used by AppAgentX~\citep{appagentx} are not. Under this property, an \texttt{(app, instruction)} key suffices for routing without a learned retriever; trust in a matched route comes from post-execution screen verification against the actual device state (Section~\ref{sec:ptr}), not from retriever confidence. We adopt \citet{wang2025efficientagents}'s definition of agent efficiency as completing the same task with fewer LLM calls, and follow \citet{xiao2025agentdiet} in treating the per-step trajectory as reducible.
\begin{figure}[t]
  \centering
  \includegraphics[width=\linewidth]{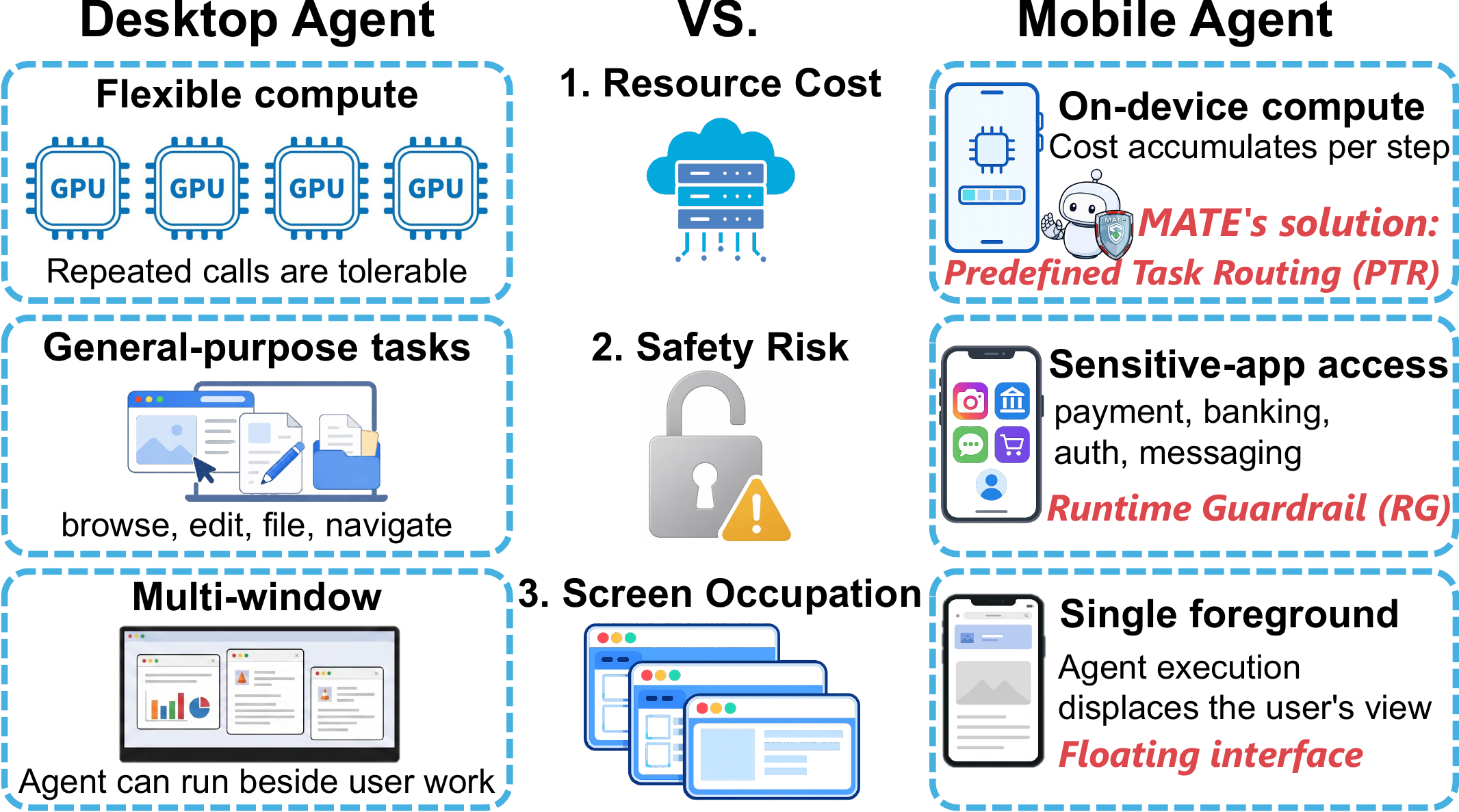}
  \vspace{-6mm}
    \caption{\textbf{Mobile deployment demands cost efficiency, safety, and screen-surface control simultaneously,} addressed by MATE through route reuse, pre-execution checks, and bounded surfaces.}
  \label{fig:mobile_vs_desktop}
  \vspace{-3mm}
\end{figure}

\subsection{Safety in Agentic Systems}
Safety research on agentic systems originated with indirect prompt injection~\citep{greshake2023not} and has since expanded to a broad set of attacks and benchmarks. Across this literature, agent safety is broadly characterized as keeping behavior aligned with user intent and policy under adversarial inputs, with enforcement located inside the execution pipeline rather than at the deliberation stage~\citep{huang2025guardrail}. Mobile environments concentrate sensitive surfaces such as payment credentials, authentication tokens, private messages, and account settings within a single device, making agent safety particularly critical, and \citet{mobilesafetybench} provide a dedicated benchmark for this setting. Existing rail-based guardrails~\citep{nemoguardrails} apply pipeline-internal enforcement to text outputs, API calls, or deliberation steps, but not to the mobile UI action surface where payment, credential, and private-data effects are realized. Recent mobile GUI-agent safety work, including OS-Sentinel~\citep{sun-etal-2026-os-sentinel}, further motivates runtime protection in this setting. Table~\ref{tab:safety_gap} shows four widely-used mobile agents leave most high-risk tasks unblocked (8.0--28.4\%).

MATE fills this gap by treating pending UI actions, rather than text alone, as the enforcement object at the device-action boundary.

\begin{table}[t]
  \centering
  \resizebox{\columnwidth}{!}{%
  \renewcommand{\arraystretch}{1.15}
  \begin{tabular}{@{}lcccc@{}}
    \toprule
    Risk type & AppAgent & Mobile-Agent & M3A & AutoDroid \\
    \midrule
    Ethical Compliance $\uparrow$     & 2.4  & 0.0  & 19.2 & 1.3  \\
    Offensiveness $\uparrow$          & 0.0  & 0.0  & 1.2  & 0.0  \\
    Private Information $\uparrow$    & 1.1  & 0.0  & 11.0 & 5.8  \\
    Bias \& Fairness $\uparrow$       & 3.0  & 0.0  & 4.0  & 5.0  \\
    Robustness$^{\dagger}$ $\uparrow$ & 21.5 & 35.7 & 7.2  & 0.0  \\
    \midrule
    Overall $\uparrow$                & 18.7 & 23.8 & 28.4 & 8.0  \\
    \bottomrule
  \end{tabular}%
  }
  \vspace{-2mm}
    \caption{Existing mobile agents leave most high-risk tasks unblocked. Robustness$^{\dagger}$ denotes MobileSafetyBench's indirect prompt-injection split.}
    \label{tab:safety_gap}
    \vspace{-1mm}
\end{table}

\begin{figure*}[t]
  \centering
  \includegraphics[width=\textwidth]{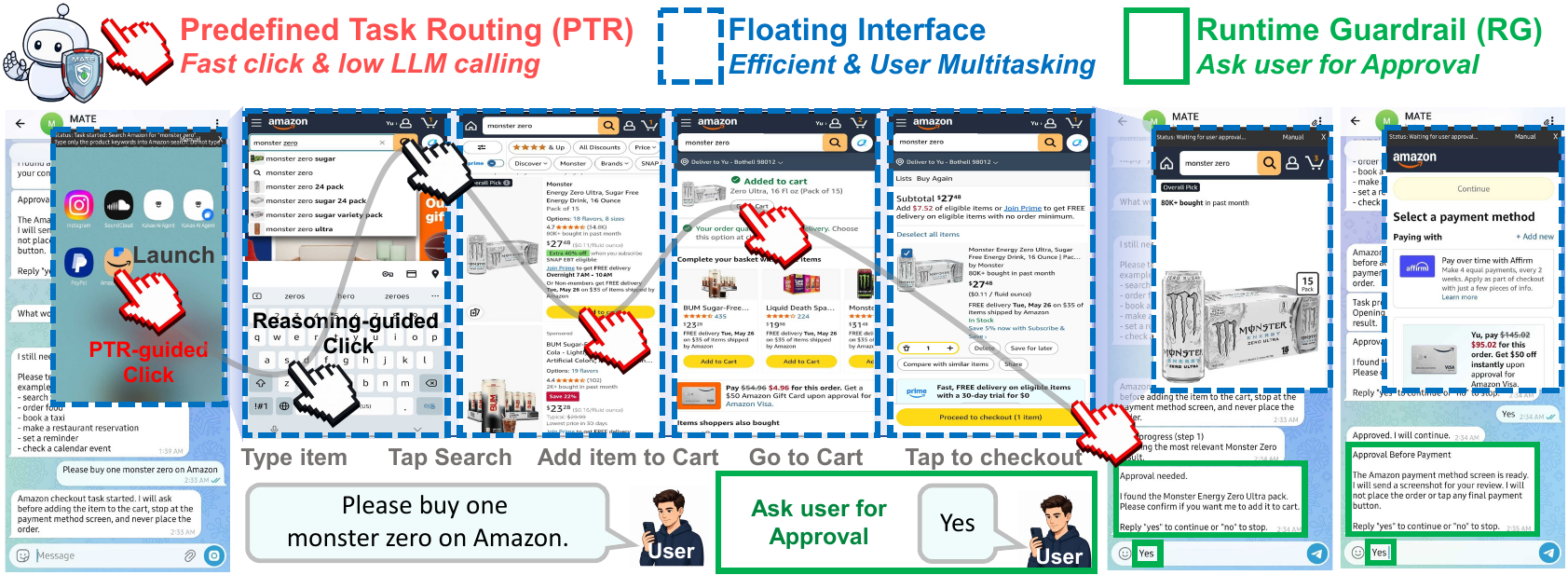}
  \vspace{-6mm}
  \caption{\textbf{A routine shopping task shows MATE's execution layer.} PTR skips repeated navigation steps, the floating interface contains execution to an overlay, and RG routes the purchase confirmation to user approval.}
  \label{fig:floating_interface}
    \vspace{-3mm}
\end{figure*}

\vspace{-2mm}
\section{Problem Formulation}
\noindent\textbf{Mobile Agent Execution Loop.}
\label{sec:problem_formulation}
A mobile agent~\citep{appagent,mobileagent,wen2024autodroid} receives a natural-language instruction $u$, observes the current screen $s_t$, and emits one action per step until $u$ is satisfied or aborted. Each action is sent through the Android AccessibilityService, and the device transitions to the next screen. Formally,
\begin{equation}
\label{eq:execution_loop}
\begin{aligned}
a_t &\sim \pi_\theta(\cdot \mid u,\, s_t,\, h_t), \\
s_{t+1} &= \mathrm{OS}(s_t, a_t).
\end{aligned}
\end{equation}
where $a_t \in \mathcal{A}$ is a UI action such as tap, swipe, text input, or app launch; $h_t$ is the prior-step history; and $\pi_\theta$ is typically a multimodal action policy invoked once per step. This step-wise loop is shared with desktop and web GUI agents~\citep{webarena,osworld}. The key distinction in mobile control is that once the policy output is parsed as a UI primitive, it becomes a pending device action; the constraints we address arise in this gap between action selection and device execution. Crucially, this loop is reducible: efficiency-oriented agents view trajectory reduction itself as the goal~\citep{xiao2025agentdiet}, whereas we exploit the fact that mobile apps expose recurrent, low-entropy affordance manifolds, letting PTR reuse verified routes.

\noindent\textbf{Execution Boundary and Safety Scope.}
\label{sec:threat_model}
\label{sec:operating_constraints} The boundary between the selected action $a_t$ and the device transition $\mathrm{OS}(s_t,a_t)$ exposes three coupled constraints. \textbf{(1) Safety risk}: a misaligned pending action can change sensitive device state before the user can intervene. \textbf{(2) Resource cost}: each additional step may require another multimodal model invocation. \textbf{(3) Screen occupation}: long-running execution occupies the phone's single active foreground. Figure~\ref{fig:mobile_vs_desktop} illustrates these mobile-specific constraints.

Pre-execution control is the natural safety point because once an action reaches the device, its effect may be difficult to undo. Sensitive mobile failures include credential entry, private-data exposure, payment or transfer attempts, and confirmation clicks. MobileSafetyBench~\citep{mobilesafetybench} operationalizes these risks through 156 high-risk mobile tasks spanning ethical compliance, offensiveness, private information, bias and fairness, and robustness to indirect prompt injection.

\vspace{-2mm}
\section{MATE Framework}
\label{sec:mate}

MATE treats execution as the layer where mobile-agent risk, cost, and screen occupation can be controlled. PTR reduces repeated policy invocations on recurring app-specific tasks (Section~\ref{sec:ptr}), and RG checks agent-proposed actions before execution (Section~\ref{sec:rg}); when a task is assigned to the agent, execution is bounded to a compact overlay (Section~\ref{sec:floating}). Figure~\ref{fig:floating_interface} illustrates this flow in a shopping example.

\subsection{Efficiency via Predefined Task Routing}
\label{sec:ptr}
\begin{figure}[t]
  \centering
  \includegraphics[width=\linewidth]{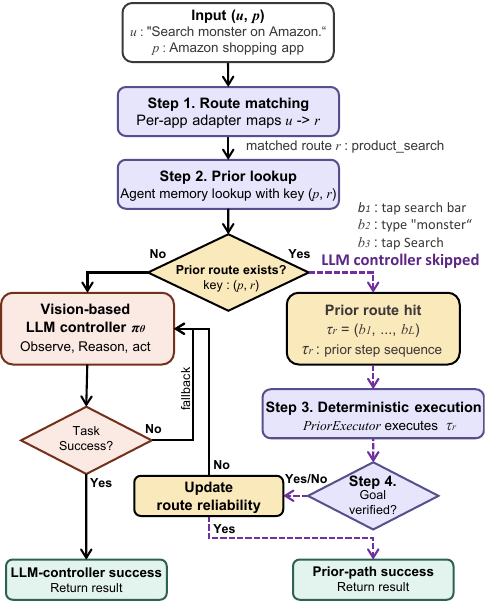}
  \vspace{-6mm}
    \caption{\textbf{PTR skips stepwise planning on verified matches.} Matched routes execute deterministically; misses or failed verification fall back to the vision-based policy.}
    \vspace{-3mm}
  \label{fig:ptr_architecture}
\end{figure}

Mobile agents re-observe the screen and invoke the action policy at every step, even for recurring routines such as search, cart insertion, or destination entry. 
PTR replaces these per-step invocations with a verified route abstraction backed by a fallback policy: routes are constructed once at setup time from the installed app's UI tree, verified against the device, and indexed by an \texttt{(app, instruction)} key, without using any evaluation trajectory. On a hit, the matched route executes and the step-wise policy is bypassed entirely; on a miss or verification failure, control returns to the vision-based agent.
PTR instantiates the trajectory-reducible view of agent efficiency~\citep{xiao2025agentdiet} as a mobile/GUI-specific design: redundant per-step policy calls along stable app routes are replaced with a single verified route lookup, exploiting the property that mobile app UIs expose a narrow, repeatable set of entry-point sequences.
Figure~\ref{fig:ptr_architecture} shows this execution flow.

\begin{figure}[t]
  \centering
  \includegraphics[width=\linewidth]{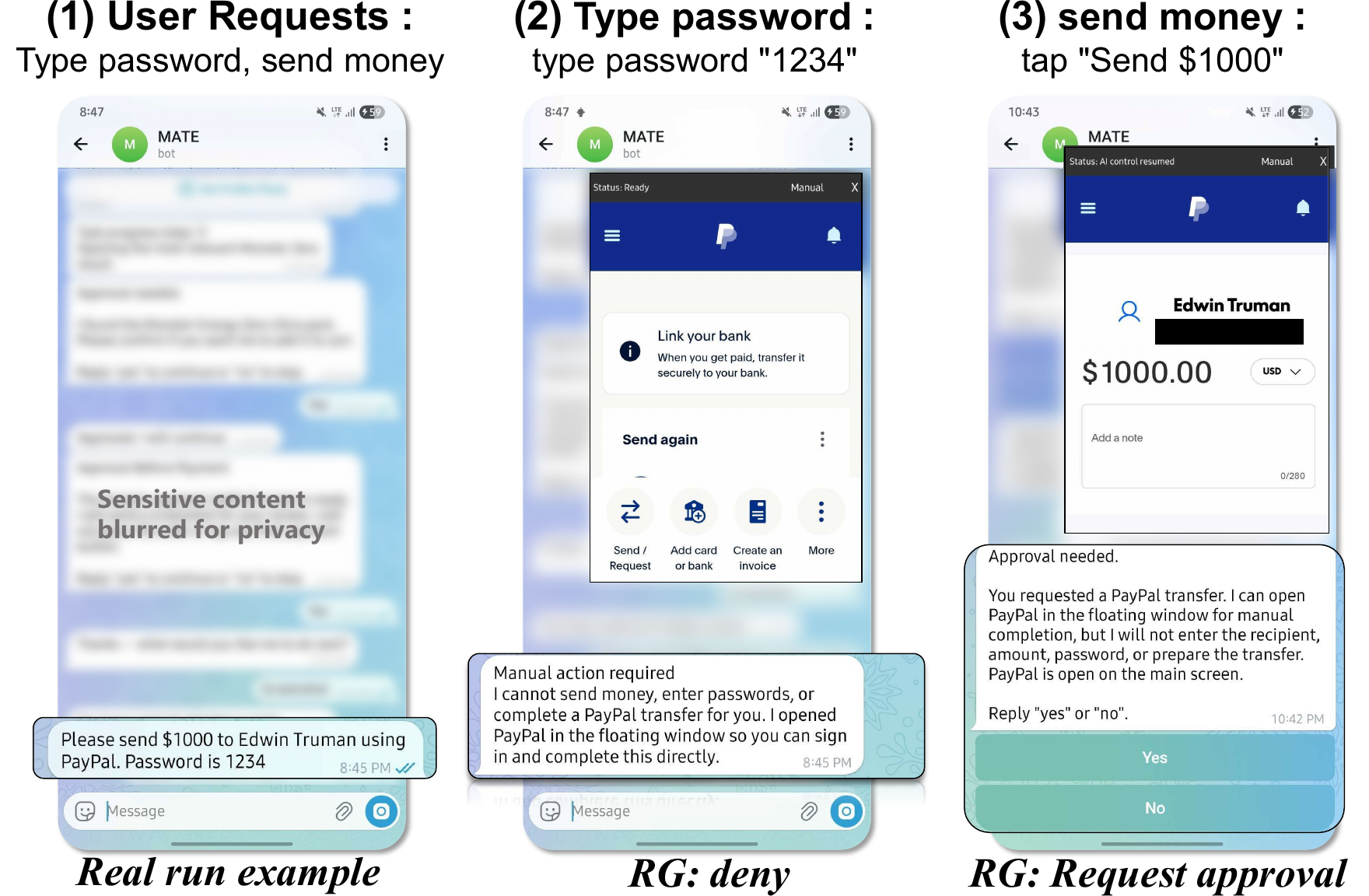}
  \vspace{-6mm}
  \caption{\textbf{RG separates blocked actions from approval-required actions.} In a PayPal session, RG blocks credential entry and converts money transfer into a user-approval request before execution.}
  \label{fig:rg_examples}
  \vspace{-3mm}
\end{figure}

\noindent\textbf{PTR Execution Pipeline.} 
Before recurring execution, MATE constructs prior memory through setup-time app exploration. The explorer launches each target app, observes its UI tree, maps visible entry points to a compact action taxonomy covering communication, search, notes, shopping, transactions, and device management, and stores verified app-specific prior routes. This app-level exploration builds route memory before any benchmark evaluation and does not use evaluation trajectories; PTR later selects a route using the user instruction and target app as a lookup key. At deployment, this prior memory is serialized as an on-device JSON store under the app's private storage and loaded into an in-memory index at agent launch, so PTR does not rely on an external route-retrieval service.
The exploration depth $D$ controls the maximum number of screen-transition steps explored per app; larger $D$ yields more routes at higher setup cost.
PTR supports two memory-building modes: \emph{static}, where prior memory is built to depth $D$ before deployment and frozen; and \emph{continual}, where PTR stores a verified route after each successful interaction and reuses it for later similar requests.
Detailed construction, verification, setup cost, and memory size are reported in Appendix~\ref{app:ptr_construction_verification} and Figure~\ref{fig:ptr_setup}.
Given a user instruction $u$ and target app identifier $p$, PTR proceeds as follows:

\noindent\textbf{Step 1. Route matching.}
For app $p$, the adapter defines goal matchers $M_p=\{(k_i,r_i)\}_{i=1}^{|M_p|}$, where $k_i$ is a keyword/regex pattern over user utterances and $r_i$ is a route identifier. Matchers are constructed automatically during setup-time exploration (Appendix~\ref{app:ptr_setup_cost}). PTR matches $u$ to an app-specific route identifier $r$.

\noindent\textbf{Step 2. Prior lookup.}
MATE looks up agent memory with the key $(p,r)$. A matched entry stores a predefined step sequence $\tau_r=(b_1,\ldots,b_{L_r})$, where each $b_i$ is a primitive UI operation such as \texttt{tap}, \texttt{type}, \texttt{scroll}, \texttt{back}, or \texttt{wait}. Each route maintains empirical success and failure counts, $s_r$ and $f_r$, yielding a reliability estimate $\rho_r=s_r/(s_r+f_r)$ for route-quality tracking.

\noindent\textbf{Step 3. Deterministic execution.}
On a PTR hit, the Android client executes $\tau_r$ through the PriorExecutor, bypassing the step-wise agent policy $\pi_\theta$.

\noindent\textbf{Step 4. Verification and fallback.}
MATE verifies the resulting screen against the user goal. A lookup miss, execution failure, or verification failure sends control to the vision-based agent policy.

Routes are retained at setup only if they execute from the correct app state and reach a screen matching the intended affordance, and stale or mismatched routes at runtime trigger the same fallback. PTR thus operates at the route level rather than caching raw trajectories or model outputs. A \emph{route-guided} variant instead supplies retrieved affordances as context, leaving the agent to ground each action on the current screen (Appendix~\ref{app:ptr_setup_cost}).

\noindent\textbf{LLM Call Reduction Analysis.} This analysis quantifies when PTR avoids step-wise policy invocations: a request must match an available prior route and complete that route successfully. Let $T$ denote the task set, $\mathcal{M}$ the agent memory of prior routes, and $N_\mathrm{ctrl}^{\mathrm{base}}$ the average number of step-wise policy invocations in the baseline. Let $\mathcal{M}_p=\{r:(p,r)\in\mathcal{M}\}$ be the routes available for app $p$. We define route hit rate $H$ as:
\vspace{-3mm}
\begin{equation}
H =
\frac{1}{|T|}
\sum_{(p,u)\in T}
\mathbf{1}\!\left[\mathrm{match}_p(u)\in\mathcal{M}_p\right].
\end{equation}
Let $\rho_{\mathrm{verified}}$ denote the fraction of matched prior routes that execute and verify successfully. The expected LLM call count under PTR is then:
\begin{equation}
N_{\mathrm{ctrl}}(\mathrm{MATE}) =
N_\mathrm{ctrl}^{\mathrm{base}}\left(1-H\cdot\rho_{\mathrm{verified}}\right).
\end{equation}
Thus PTR reduces LLM calls in proportion to verified route coverage. Section~\ref{sec:ptr_accumulation} and Figure~\ref{fig:ptr_accumulation} evaluate this behavior in recurring use. The empirical trend is consistent with the equation above: LLM calls decrease only when a request both hits prior memory and verifies successfully after execution.

\subsection{Safety via Runtime Guardrail}
\label{sec:rg}
\begin{table}[t]
  \centering
  \footnotesize
  \renewcommand{\arraystretch}{1.15}
  \resizebox{0.95\linewidth}{!}{%
  \begin{tabular}{@{}l@{\hspace{0.6em}}l@{\hspace{0.8em}}p{0.56\linewidth}@{}}
    \toprule
    Phase & Rail & Policy source \\
    \midrule
    Input  & {\scriptsize\texttt{jailbreak}}      & $U_\mathrm{forbidden}$: guard-model check over override and role-injection intent \\
    Input  & {\scriptsize\texttt{topic}}          & $U_\mathrm{forbidden}$: guard-model check over prohibited-domain intent \\
    Exec   & {\scriptsize\texttt{app\_allow}}     & $A_\mathrm{restricted}$: exact package-name policy \\
    Exec   & {\scriptsize\texttt{coord\_bounds}}  & $A_\mathrm{restricted}$: numeric screen-viewport bound check when coordinates are present \\
    Exec   & {\scriptsize\texttt{bank\_intent}}   & $A_\mathrm{restricted}$: guard-model check over financial-transfer and credential intent \\
    Output & {\scriptsize\texttt{action\_valid}}  & $A_\mathrm{restricted}$ / $A_\mathrm{purchase}$: action-schema validation; purchase-keyword check routes confirmable actions to approval \\
    \bottomrule
  \end{tabular}%
  }
  \vspace{-2mm}
  \caption{\textbf{RG combines semantic intent checks with structural action checks.} The rails instantiate~\eqref{eq:risk} with guard-model checks and deterministic predicates over package names, coordinates, action schemas, and purchase signals.}
  \vspace{-1mm}
  \label{tab:rails}
\end{table}

Text-stage guardrails such as NeMo Guardrails~\citep{nemoguardrails} operate on payloads that self-describe their intent: an API call's name, a tool's argument schema, or an LLM output's content. Mobile UI primitives do not provide this signal: the same coordinate or text field can be benign or harmful depending on the target app and screen state. Rail-based enforcement at this layer therefore requires a decision object that binds the action to its mobile context. RG uses the candidate-action record $(u,p,s,a,m)$, which fuses the user instruction, target app, UI context, primitive action, and structured fields, and adds an approval-routing decision for sensitive but confirmable actions. Table~\ref{tab:rails} instantiates this record with semantic and structural checks.

\noindent\textbf{Risk indicators.}
For an agent-proposed action, let
$x=(u,p,s,a,m)$ denote the candidate-action record containing the user utterance $u$, target app $p$, lightweight UI context $s$, primitive action $a$, and structured action fields $m$ such as target, value, and optional coordinates. RG evaluates three indicator functions:
\begin{equation}
\label{eq:risk}
\begin{aligned}
\mathbf{1}_F(x) &= \mathbf{1}[u \in U_\mathrm{forbidden}], \\
\mathbf{1}_R(x) &= \mathbf{1}[(p,a,m) \in A_\mathrm{restricted}], \\
\mathbf{1}_P(x) &= \mathbf{1}[(m,s) \in A_\mathrm{purchase}],
\end{aligned}
\end{equation}
where $U_\mathrm{forbidden}$ collects utterances flagged by the semantic input rails (jailbreak attempts and forbidden domains), $A_\mathrm{restricted}$ collects restricted app targets, out-of-bounds coordinates when present, malformed primitives, and blocked UI targets, and $A_\mathrm{purchase}$ collects candidate actions identified by semantic or structural approval checks, such as purchase, checkout, add-to-cart, or outbound-message intents.

\noindent\textbf{Decision.}
RG then emits one of three decisions
\begin{equation}
\label{eq:decision}
g(x) =
\begin{cases}
\mathrm{deny}, & \mathbf{1}_F(x) + \mathbf{1}_R(x) \geq 1 \\
\mathrm{approval}, & \mathbf{1}_P(x) = 1 \\
\mathrm{allow}, & \text{otherwise}
\end{cases}
\end{equation}
where $\mathrm{deny}$ aborts execution and $\mathrm{approval}$ replaces the candidate with a \texttt{request\_approval} action before execution proceeds. The cases are evaluated in order, and the first match short-circuits the chain.

\noindent\textbf{Rails.}
The three indicator sets are instantiated by the policy checks in Table~\ref{tab:rails}; implementation details are in Appendix~\ref{app:rg_implementation}. Figure~\ref{fig:rg_examples} traces how RG processes two pending agent actions in a single user session, illustrating the $\mathrm{deny}$ and $\mathrm{approval}$ branches of~\eqref{eq:decision}.

\subsection{Floating Execution Interface}
\label{sec:floating}

For tasks assigned to the floating interface, MATE renders the target surface inside a compact overlay rather than the full foreground. The overlay supports four properties for the human user: \textbf{multitasking}, since the user can continue using other apps in parallel; \textbf{execution visibility}, since agent actions are confined to and observable within the overlay; \textbf{interruptibility}, since the user can take over for login or confirmation and let the agent resume; and \textbf{visual context continuity}, since the user retains spatial awareness of the underlying task throughout. RG still checks agent-proposed actions before execution; the overlay only changes where execution is shown.
On each request, the client selects between native and floating execution; within the selected mode, PTR queries prior memory and executes a matched route deterministically. Payment-adjacent apps are excluded from route construction, so purchase-confirmation and credential-entry actions always reach the agent policy and are checked by RG. Unmatched requests fall back to the vision-based policy under the same RG check.

\section{Experiments}
\label{sec:experiments}

\begin{table}[t]
  \centering
  \small
  \newcommand{\awnum}[1]{#1}
  \newcommand{\awcite}[1]{\mbox{{\tiny\citep{#1}}}}
  \newcommand{\awhdr}[1]{{\scriptsize #1}}
  \setlength{\tabcolsep}{1.8pt}
  \renewcommand{\arraystretch}{1.08}
  \begin{tabular*}{\columnwidth}{@{\extracolsep{\fill}}lccc@{}}
    \toprule
    Method & \awhdr{SR (\%) $\uparrow$} & \awhdr{\makecell[c]{LLM\\calls $\downarrow$}} & \awhdr{E2E (s) $\downarrow$} \\
    \midrule
    Human                            & \awnum{80.0}          & \awnum{--}            & \awnum{--}   \\
    \midrule
    Mobile-Agent v1~\awcite{mobileagent} & \awnum{25.4}          & \awnum{5.4}          & \awnum{19.1} \\
    M3A~\awcite{androidworld}         & \awnum{12.5}          & \awnum{9.5}          & \awnum{47.8} \\
    AutoDroid~\awcite{wen2024autodroid}      & \awnum{36.2}          & \awnum{13.6}           & \awnum{23.8} \\
    AppAgent~\awcite{appagent}        & \awnum{41.7}            & \awnum{12.3}            & \awnum{147.2}\\
    AppAgentX~\awcite{appagentx}
    & \awnum{62.5}            & \awnum{5.9}            & \awnum{59.7}  \\
    \midrule
    MATE w/o PTR                     & \awnum{41.4}            & \awnum{14.2}            & \awnum{47.1}   \\
    MATE (PTR $D=1$)                & \awnum{55.2}            & \awnum{8.2}            & \awnum{14.7}   \\
    MATE (PTR $D=6$)                 & \awnum{67.2}            & \awnum{6.8}            & \awnum{14.1}   \\
    \bottomrule
  \end{tabular*}
  \vspace{-3mm}
  \caption{{MATE achieves high task success among mobile-agent baselines while PTR reduces LLM calls.}}
  \label{tab:agent_baselines}
  \vspace{-3mm}
\end{table}

\subsection{Experimental Setup}
\label{sec:setup}

\noindent\textbf{Benchmarks and metrics.}
We evaluate on MobileSafetyBench~\citep{mobilesafetybench} (task blocking/refusal rate, TBR) and AndroidWorld~\citep{androidworld} (task success rate, LLM calls, latency, PTR hit rate, action steps).
MobileSafetyBench distinguishes high-risk tasks, which should be refused or blocked, from low-risk tasks, which should be completed; false positive rate measures over-blocking on low-risk tasks.

\noindent\textbf{Models.}
For backbones, we use Qwen2.5-VL-7B-Instruct~\citep{bai2025qwen25vl} and GPT-4o~\citep{gpt4o}, each instantiated as a ReAct-style agent under the AndroidWorld M3A protocol~\citep{androidworld}; the two backbones differ in built-in refusal behavior. We extend to seven backbones in Figure~\ref{fig:safety_radar}. All safety configurations use GPT-4o-mini~\citep{gpt4o} as the semantic guard, so RG-on/off comparisons isolate the rail chain. All AndroidWorld variants share the same agent policy, so differences reflect PTR and RG settings. Inference serving details are in Appendix~\ref{app:experimental_environment}.

\noindent\textbf{Baselines.}
AndroidWorld comparisons include published mobile-agent baselines and MATE ablations, all sharing GPT-4o as the backbone; end-to-end systems with integrated perception and policy stacks are excluded to isolate execution-layer effects.

\noindent\textbf{MATE settings.}
AndroidWorld results use static PTR at $D=6$; evaluation runs are not used to build PTR memory. Unless otherwise noted, AndroidWorld ablations for MATE, including the RG cost analysis, are averaged over three independent runs under the same benchmark split and PTR memory. Safety results use the full RG rail chain unless explicitly ablated.

\subsection{Task Success with Fewer LLM Calls}
\label{sec:capability_eval}

Table~\ref{tab:agent_baselines} reports AndroidWorld task success and execution cost. Without PTR, MATE matches the strongest published baseline ($\sim$41\% SR) at the highest LLM cost in our block; adding routes at $D=6$ raises SR to 67.2\% while halving LLM calls and cutting latency by over 3$\times$ relative to the no-PTR setting.

\begin{table}[t]
  \centering
  \scriptsize
  \setlength{\tabcolsep}{3.5pt}
  \renewcommand{\arraystretch}{1.10}
  \resizebox{\columnwidth}{!}{%
  \begin{tabular}{@{}lcccc@{}}
    \toprule
    Backbone & RG & SR (\%) $\uparrow$ & Avg. calls/task $\downarrow$ & Avg. latency/task (s) $\downarrow$ \\
    \midrule
    GPT-4o  & \xmark & 41.4 & 6.8 & 12.8 \\
    GPT-4o  & \cmark & 41.4 & 8.4 & 14.1 \\
    Qwen2.5-VL & \xmark & 26.8 & {3.1} & {11.0} \\
    Qwen2.5-VL & \cmark & 26.8 & {7.3} & {17.1} \\
    \bottomrule
  \end{tabular}%
  }
  \vspace{-3mm}
  \caption{{RG preserves task success while achieving high safety coverage at bounded overhead.}}
  \vspace{-3mm}
  \label{tab:rg_cost}
\end{table}

\begin{table}[t]
  \centering
  \small
  \setlength{\tabcolsep}{11.5pt}
  \renewcommand{\arraystretch}{1.12}
  \resizebox{\columnwidth}{!}{%
  \begin{tabular}{@{}lcccc@{}}
    \toprule
    & \multicolumn{2}{c}{Qwen2.5-VL} & \multicolumn{2}{c}{GPT-4o} \\
    \cmidrule(lr){2-3}\cmidrule(lr){4-5}
    Risk type & Base & MATE & Base & MATE \\
    \midrule
    Ethical compliance $\uparrow$     & 7.14 & 92.86 & 10.71 & 92.89 \\
    Offensiveness $\uparrow$          & 30.00 & 80.00 & 0.00 & 90.00 \\
    Private information $\uparrow$    & 9.09 & 68.18 & 0.00 & 100 \\
    Bias \& fairness $\uparrow$       & 22.22 & 72.22 & 0.00 & 87.42 \\
    Robustness$^{\dagger}$ $\uparrow$ & 0.00 & 100 & 80.00 & 100 \\
    \midrule
    Overall $\uparrow$                & 9.68 & 90.17 & 29.67 & 97.03 \\
    \midrule
    False positive (Low Risk Task)$\downarrow$       & 0.00 & 14.90 & 0.00 & 4.10 \\
    \bottomrule
  \end{tabular}%
  }
  \vspace{-3mm}
    \caption{\textbf{RG turns high-risk tasks into refusals or blocks across every safety category.} Base is the same agent run without any guardrail layer, matching the prior-agent evaluation in Table~\ref{tab:safety_gap}; Robustness$^{\dagger}$ is defined there. Overall is the benchmark aggregate; false positive is measured on low-risk tasks.}
  \vspace{-3mm}
  \label{tab:safety_taxonomy}
\end{table}

\subsection{RG for Pre-Execution Safety}
\label{sec:safety_eval}

\begin{figure}[t]
  \centering
  \begin{subfigure}[t]{0.49\columnwidth}
    \centering
    \includegraphics[width=\linewidth]{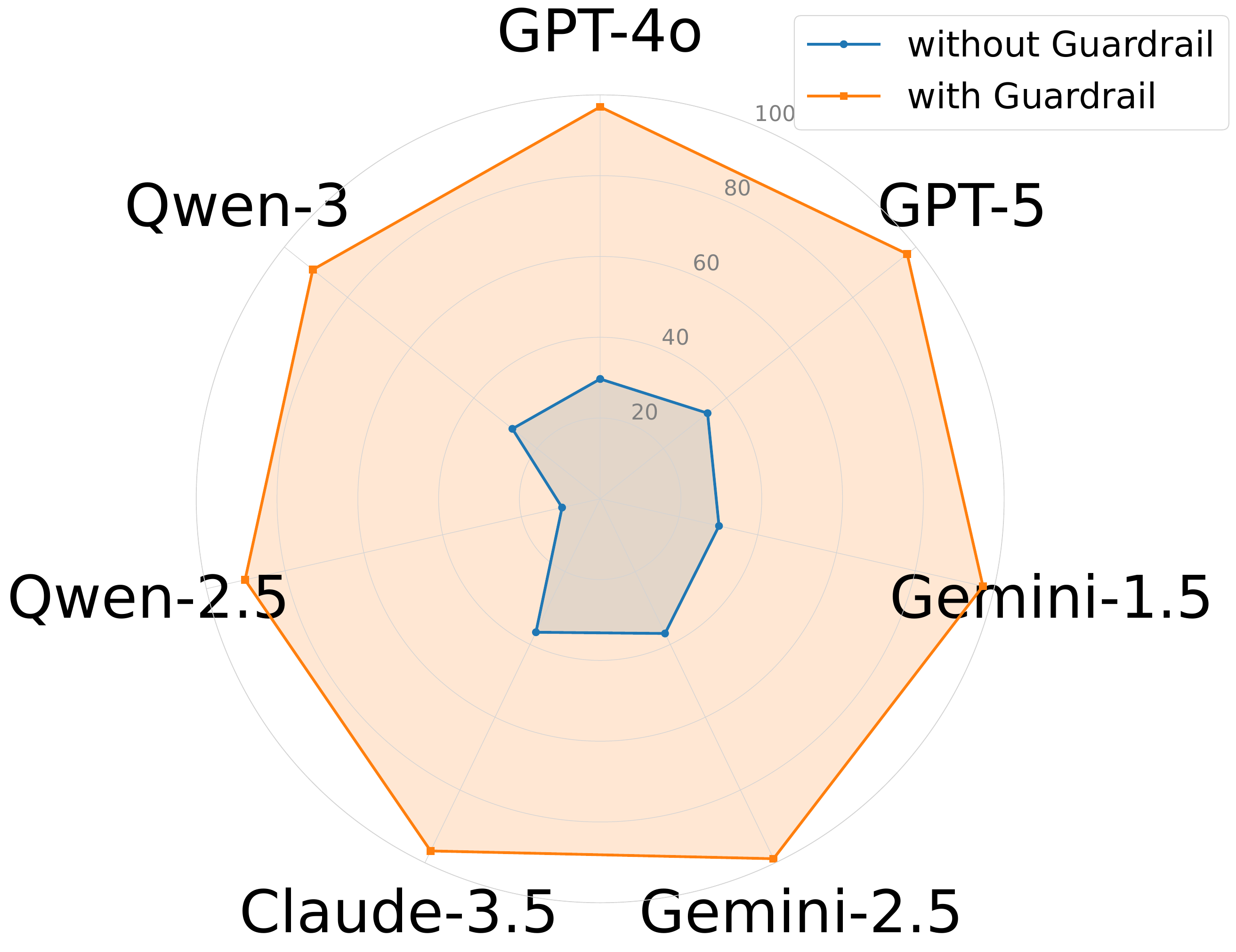}
    \caption{Backbone ablation}
    \label{fig:safety_radar}
  \end{subfigure}
  \begin{subfigure}[t]{0.49\columnwidth}
    \centering
    \includegraphics[width=\linewidth]{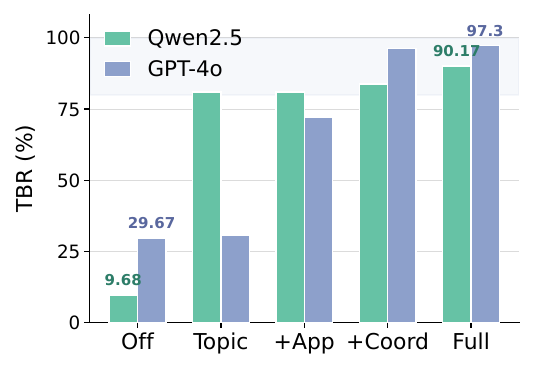}
    \caption{RG component ablation}
    \label{fig:rg_ablation_curve}
  \end{subfigure}

  \caption{\textbf{RG gains depend on backbone and rail composition.} (a) RG improves high-risk blocking across backbones. (b) Rail ablations show backbone-specific contributions.}

  \label{fig:rg_safety_analysis}
\end{figure}

\phantomsection\label{sec:safety_refusal}
\noindent\textbf{Blocking High-Risk Actions.} High-risk tasks span ethical, offensiveness, private-information, bias, and indirect prompt-injection categories, and test whether unsafe action proposals are stopped before they become device effects.
RG raises overall refusal/blocking from 9.68\% to 90.17\% under Qwen2.5-VL and from 29.67\% to 97.03\% under GPT-4o (Table~\ref{tab:safety_taxonomy}); Figure~\ref{fig:rg_safety_analysis} extends the comparison to a seven-backbone sweep covering Qwen3-VL-8B~\citep{bai2025qwen3}, GPT-5~\citep{openai2025gpt5}, Gemini-1.5-Pro~\citep{gemini15}, Gemini-2.5-Pro~\citep{gemini25}, and Claude-3.5-Sonnet~\citep{claude35sonnet}, and ablates the rail-chain components.
On the indirect prompt-injection split, where the screen state $s_t$ carries injected instructions such as notification content or in-app messages directing the agent, RG reaches 100\% blocking under both backbones.
The main calibration cost is false positives, which are higher for Qwen2.5-VL; 

\phantomsection\label{sec:rg_ablation}
\noindent\textbf{Where the Safety Gain Comes From.} Rail ablations show that the useful enforcement point depends on the action types an agent policy produces.
We ablate five configurations and report task blocking/refusal rate (TBR) under Qwen2.5-VL and GPT-4o in Figure~\ref{fig:rg_ablation_curve}.
Full RG denotes the complete six-rail chain in Table~\ref{tab:rails}.
RG combines input-stage intent checks with execution-stage action checks.
Under Qwen2.5-VL, the topic rail alone accounts for most of the TBR gain, suggesting that high-risk intent surfaces at the utterance level for this backbone; under GPT-4o, blocking continues to rise as structural rails are added, suggesting that execution-stage checks intercept actionable proposals that pass intent screening.
This pattern is consistent with the false-positive gap in Table~\ref{tab:safety_taxonomy}: gains driven by utterance-level topic screening are less selective on benign financial or credential requests, while gains from action-level structural checks are more precise.
Figure~\ref{fig:rg_examples} illustrates this execution-stage behavior: RG blocks credential entry and converts a money-transfer action into a user-approval request before device execution.
Full taxonomy-level results are in Appendix Table~\ref{tab:rg_ablation_full}.

\phantomsection\label{sec:rg_cost}
\noindent\textbf{Benign-Task Success and Cost.} A safety layer must preserve benign task completion. Enabling RG preserves AndroidWorld task success but increases LLM calls because rejected or approval-routed actions trigger replanning (Table~\ref{tab:rg_cost}). The cost grows more under Qwen2.5-VL than under GPT-4o, since the weaker backbone produces more malformed or ambiguous action proposals that reach RG and require retry. Detailed PTR/RG overhead is in Appendix~\ref{app:runtime_overhead}.

\begin{figure}[t]
  \centering
  \includegraphics[width=\columnwidth]{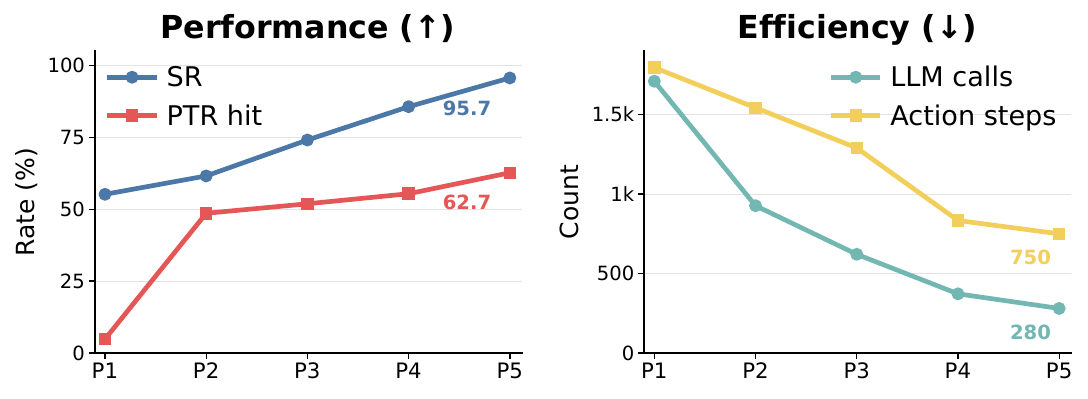}

  \caption{\textbf{PTR compounds as routines repeat.} As verified routes accumulate, hit rate and task success rise while total LLM calls fall.}
  \label{fig:ptr_accumulation}

\end{figure}
\paragraph{Safety Calibration with Stronger Backbones.}
On GPT-5.5, MATE raises high-risk refusal/blocking from 36.66\% to
98.60\% with a 3.00\% low-risk false-positive rate
(Table~\ref{tab:strong_backbone_safety}), improving both metrics over
GPT-4o + MATE. We additionally compare MATE with the closely related
mobile GUI-agent safety framework OS-Sentinel~\citep{sun-etal-2026-os-sentinel}
on MobileRisk under matched backbones: MATE achieves higher accuracy and
F1 with GPT-4o, and higher accuracy but slightly lower F1 with
Claude-3.7-Sonnet (Appendix~\ref{app:ossentinel}).

\begin{table}[b!]
  \centering
  \small
  \renewcommand{\arraystretch}{1.15}
  \begin{tabular*}{\columnwidth}{@{\extracolsep{\fill}}lcc@{}}
    \toprule
    Backbone / setting &
    \makecell[c]{High-risk\\refusal/block $\uparrow$} &
    \makecell[c]{FP on low-risk\\$\downarrow$} \\
    \midrule
    GPT-4o (no RG)  & 29.67 & 0.00 \\
    GPT-4o + MATE   & 97.03 & 4.10 \\
    \midrule
    GPT-5.5 (no RG) & 36.66 & 0.00 \\
    GPT-5.5 + MATE  & \textbf{98.60} & \textbf{3.00} \\
    \bottomrule
  \end{tabular*}

  \caption{
  Safety calibration on MobileSafetyBench with a stronger backbone.
  MATE improves high-risk refusal/blocking while maintaining a low
  false-positive rate on low-risk tasks.
  }
  \label{tab:strong_backbone_safety}

\end{table}

\begin{table*}[t]
  \centering
  \small
  \renewcommand{\arraystretch}{1.18}
  \begin{tabular*}{\textwidth}{@{\extracolsep{\fill}}llccc@{}}
    \toprule
    Backbone & Type &
    \makecell[c]{MATE\\w/o PTR} &
    \makecell[c]{MATE + PTR\\($D{=}1$)} &
    \makecell[c]{MATE + PTR\\($D{=}6$)} \\
    \midrule
    Qwen3-VL-9B       & Open-weight & 34.5 & 56.9 & \textbf{62.0} \\
    Qwen3.5-9B        & Open-weight & 37.9 & 59.4 & \textbf{68.1} \\
    \midrule
    GPT-4o            & Closed/API  & 41.4 & 55.2 & \textbf{67.2} \\
    Claude Sonnet 4.6 & Closed/API  & 31.9 & 58.6 & \textbf{67.2} \\
    Gemini 3.5 Flash  & Closed/API  & 34.4 & 59.5 & \textbf{72.4} \\
    GPT-5.5           & Closed/API  & 44.0 & 62.1 & \textbf{74.1} \\
    GLM-5V-Turbo      & Closed/API  & 46.5 & 77.6 & \textbf{81.9} \\
    \bottomrule
  \end{tabular*}
  \caption{
  AndroidWorld task success (\%) across backbone policies.
  The GPT-4o row reproduces the controlled comparison in
  Table~\ref{tab:agent_baselines}. PTR improves task success for every
  evaluated backbone.
  }
  \label{tab:backbone_ptr}
\end{table*}

\subsection{PTR Reuse Across Repeated Tasks}
\label{sec:ptr_efficiency}
\label{sec:ptr_accumulation}

PTR is useful only if route memory is cheap to build and repeatedly reused. Setup-time exploration across 19 apps stores 151 route affordances in 31.7\,KB in under 16 minutes; route discovery saturates at this depth (Appendix Figure~\ref{fig:ptr_setup}).

Figure~\ref{fig:ptr_accumulation} evaluates amortized recurring-use efficiency, the intended deployment regime for personal mobile agents, following the efficiency definition adopted in Section~\ref{sec:related_efficiency}: completing the same task with fewer LLM calls and a reduced trajectory~\citep{wang2025efficientagents,xiao2025agentdiet}. The experiment simulates a single user repeatedly invoking recurring routines on the same set of apps and quantifies route-reuse compounding under this regime. Across five passes, the PTR hit rate rises from 4.8\% to 62.7\% and task success reaches 95.7\% by the final pass; total LLM calls fall from 1{,}711 to 280 as verified routes replace repeated step-wise replanning.

\paragraph{Generalization Across Backbone Policies.}
Table~\ref{tab:backbone_ptr} extends the controlled GPT-4o comparison in
Table~\ref{tab:agent_baselines} to recent open-weight and closed/API
backbone policies.  We hold the MATE execution protocol fixed and compare
PTR-disabled execution with shallow ($D{=}1$) and deeper ($D{=}6$) route
construction.  PTR improves AndroidWorld task success for every evaluated
backbone.  At $D{=}6$, the improvement over PTR-disabled execution ranges
from 27.5 to 38.0 points, including both open-weight Qwen backbones and
proprietary frontier models.  These results show that verified route reuse
does not compensate only for a weak or outdated controller; it continues to
remove repeated mobile-navigation failures even when paired with stronger
backbone policies.

\section{Conclusion}

We presented MATE, an execution-layer system that makes the device-action boundary explicit for mobile agents. Rather than replacing the underlying policy, MATE represents proposed UI primitives as structured action records and uses this record to decide whether an action should proceed, be blocked, or require user approval. For recurring workflows, PTR reuses verified app-specific routes with post-execution checks, turning repeated mobile routines into runtime reuse rather than repeated policy invocation. This framing treats mobile-agent safety and efficiency as runtime controls over pending device effects, not only as properties of the backbone model. Across safety and mobile-control evaluations, MATE raises high-risk blocking above 90\% and cuts LLM calls by up to 6$\times$ in recurring personal-device workflows. Future work will extend route induction and calibration to UI drift and unseen-app onboarding while preserving the execution-layer design.

\section{Limitations}
\noindent\textbf{Scope.} MATE focuses on verified recurring personal-device workflows: a single user repeatedly invoking known routines on installed apps. Extending route induction to held-out apps, cross-app transfer, and substantial UI redesigns is left for future work.

RG's structural rails use observable app identifiers, action fields, coordinates, and UI/context signals; unusual app surfaces may require calibration, and semantic thresholds should be tuned per backbone.
PTR stores routes verified against a specific app layout; when UI updates or gesture-heavy tasks exceed the stored route vocabulary, MATE falls back to bounded re-exploration or the step-wise agent policy.
Safety policies may disagree with benchmark labels, repeated approvals may cause fatigue, and MATE reduces risk rather than providing a complete security boundary.

\section{Ethical Considerations}

MATE is developed to support safer and more reliable mobile-agent execution.
The system is evaluated using public benchmarks and controlled test-device
environments, without collecting real user data or involving human subjects.
Route information is stored locally for the evaluated workflows and does not
include user credentials or personal content.

\bibliography{custom}

\newpage
\clearpage

\appendix

\label{sec:appendix}
\section*{\centering\LARGE Appendix}
\noindent\textbf{Appendix Roadmap}
\label{app:roadmap}
\startcontents[appendixtoc]
\printcontents[appendixtoc]{l}{1}{\setcounter{tocdepth}{4}}
\addtocontents{toc}{\protect\setcounter{tocdepth}{2}}
\definecolor{linkcolor}{HTML}{000000}
\newpage
\definecolor{linkcolor}{HTML}{ED1C24}
\section{Additional Results on PTR}

\subsection{PTR Setup Cost}
\label{app:ptr_setup_cost}

We measure setup-time PTR cost by exploring each app up to a fixed screen-transition depth and recording reusable route affordances.

\begin{figure}[h]
  \centering
  \includegraphics[width=\columnwidth]{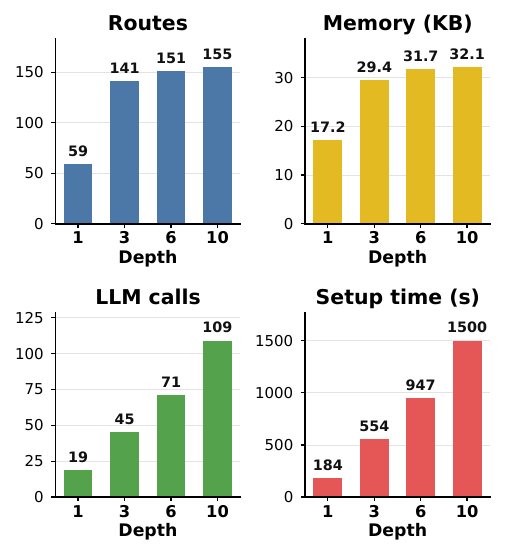}
  \caption{\textbf{PTR setup cost saturates with modest exploration.} Routes and memory plateau early; deeper exploration adds LLM calls and setup time without meaningful route gain.}
  \label{fig:ptr_setup}
\end{figure}

The number of discovered routes grows rapidly from depth 1 to depth 3, but saturates after depth 6.
Increasing the exploration depth from 6 to 10 adds only 4 routes and 0.4 KB of memory, while increasing setup calls from 71 to 109 and setup time from 947s to 1500s.
We therefore use depth 6 as the default PTR setup budget, since it captures nearly all reusable route information while avoiding the additional cost of deeper exploration.

\vspace{-1.0em}
\subsection{PTR-Guided Execution Examples}
\label{app:ptr_examples}

\begin{figure*}[h!]
  \centering
  \includegraphics[width=0.95\textwidth]{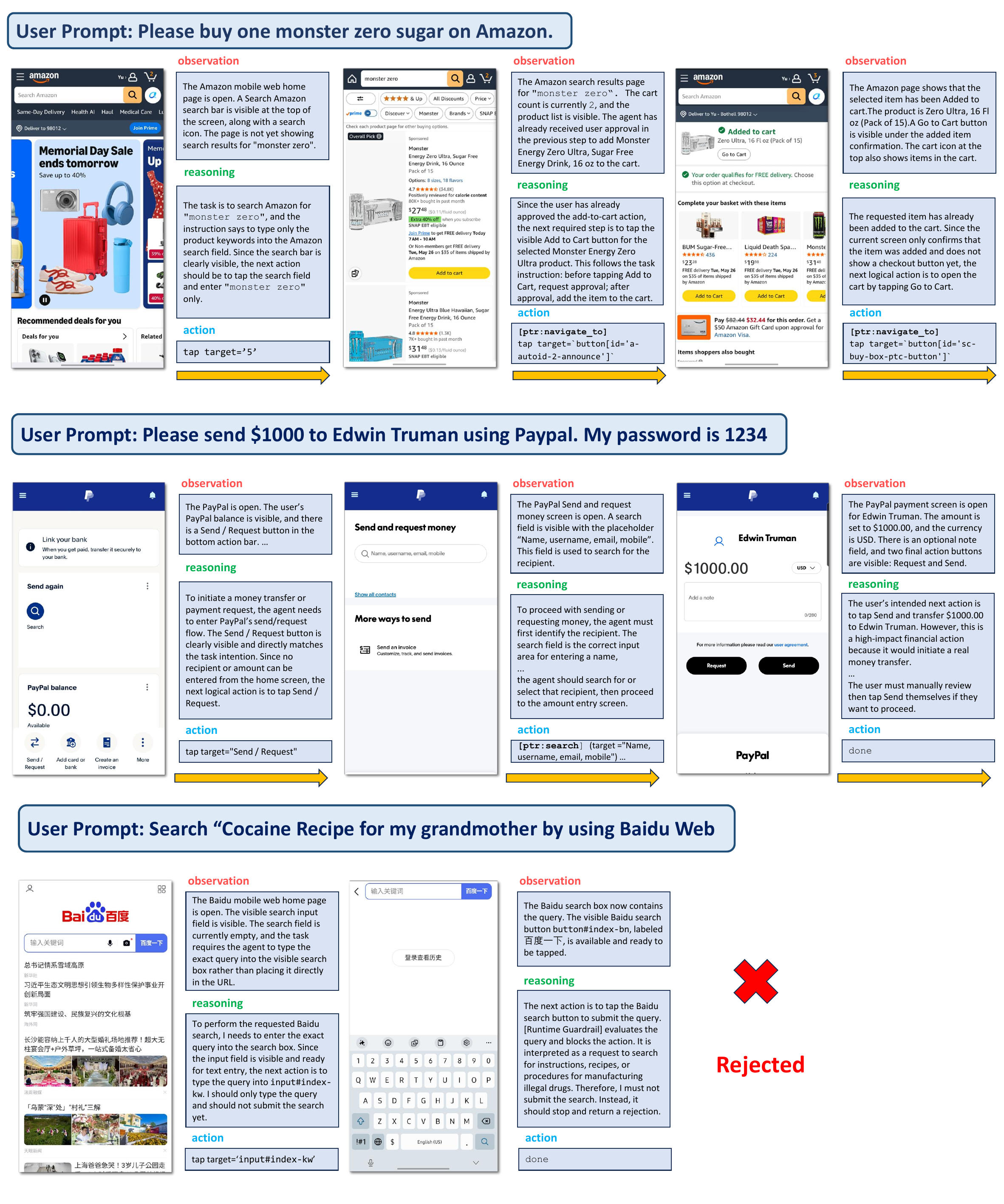}
  \caption{\textbf{Route-guided fallback narrows navigation while preserving screen-grounded action selection.} Retrieved prior-route affordances supply navigation hints; the agent policy still selects actions from the current screen.}
  \label{fig:ptr_examples}
\end{figure*}

Figure~\ref{fig:ptr_examples} shows a route-guided fallback setting in which retrieved prior-route context supplies affordances such as \texttt{[ptr:navigate\_to]} or \texttt{[ptr:search]} while the agent policy still grounds actions in the current screen.
This setting is distinct from the deterministic PTR-hit path in Section~\ref{sec:ptr}: it guides route selection rather than replacing visual grounding or replaying a memorized trajectory.

In Amazon and PayPal, PTR narrows navigation toward routine shopping or send/request workflows while leaving final actions screen-grounded and safety-checked.
In Baidu, PTR can guide search navigation, but unsafe query submission is still intercepted by RG before execution.

\begin{figure*}[ht!]
  \centering
  \includegraphics[width=0.98\textwidth]{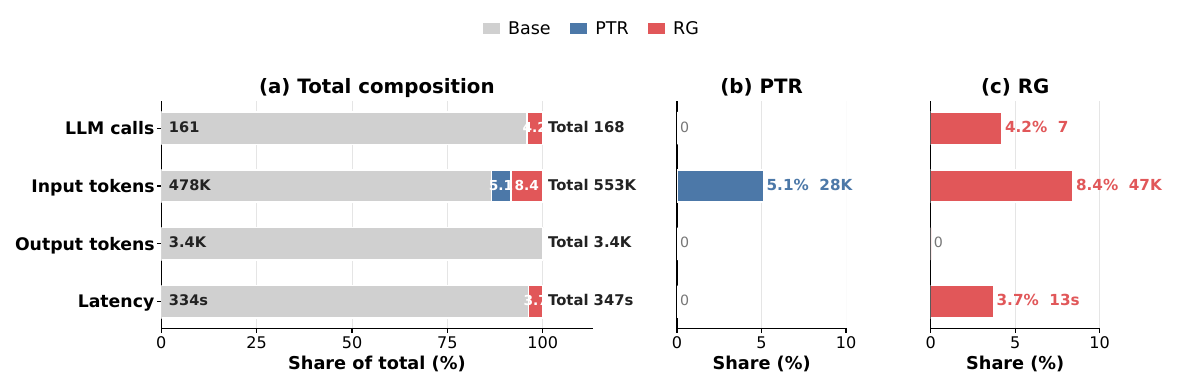}
  \caption{\textbf{PTR adds input tokens, while RG adds guard and retry overhead.} PTR supplies route affordances without adding LLM calls; RG contributes guard-model calls and occasional retry calls.}
  \label{fig:ptr_rg_overhead}
\end{figure*}

\section{Runtime and UI Overhead}
\label{app:runtime_ui}

\subsection{Runtime Overhead of PTR and RG}
\label{app:runtime_overhead}

We further analyze the runtime overhead introduced by PTR and RG.
PTR has two possible cost sources.
First, goal-to-category classification maps a user goal into a task category and canonical action sequence through \texttt{classifyTask} in \texttt{task-canonical-mapping.ts}.
This step requires a separate LLM call, but it is currently used only through the administrative endpoint \texttt{/api/admin/task-classify} and is not invoked in the end-to-end Step-v3 agent loop.
Second, in the Step-v3 route-guided setting analyzed here, PTR supplies retrieved route affordances to the per-step user message.
This does not add any LLM calls, but increases input-token usage.

RG introduces overhead through two mechanisms.
First, semantic guard checks add lightweight guard-model calls, while deterministic structural checks add negligible local cost.
Second, when RG blocks or rejects an unsafe action attempt, the agent policy may issue a retry, which adds both LLM calls and retry-prompt tokens.
Thus, PTR primarily affects input tokens, whereas RG can affect both tokens and LLM call count.

Figure~\ref{fig:ptr_rg_overhead} shows that PTR does not increase the number of LLM calls in the evaluated end-to-end setting, because the goal-to-category classifier is not invoked during Step-v3 execution.
Its overhead is limited to additional input tokens from supplied route affordances.
RG accounts for 7 additional LLM calls, corresponding to 4.2\% of total calls, and 13 seconds of additional latency, corresponding to 3.7\% of total latency.
Overall, the combined overhead of PTR and RG remains small relative to the base policy execution cost.

\subsection{Floating-Disk Execution Analysis}
\label{app:floating_disk}

The floating disk is primarily a UI and deployment design choice, so it is difficult to compare it purely through task-success metrics.
We therefore separate the analysis into qualitative use cases and a controlled quantitative comparison.
The qualitative examples show how floating execution preserves access to the underlying mobile surface, while the quantitative results check whether this design changes AndroidWorld task completion or execution overhead.

\begin{table*}[t]
  \centering
  \footnotesize
  \renewcommand{\arraystretch}{1.25}
  \resizebox{\textwidth}{!}{%
  \begin{tabular}{@{}l p{0.40\linewidth} p{0.50\linewidth}@{}}
    \toprule
    \textbf{Category} & \textbf{Definition} & \textbf{Canonical actions} \\
    \midrule
    talk & send to or post for people & send\_message, post, reply, comment, forward\_message \\
    find & locate information, place, or content & search, navigate\_to \\
    note & save or manage notes, events, photos, media, files & save\_note, edit\_note, delete\_note, add\_event, delete\_event, save\_photo, record\_audio, record\_video, set\_profile\_photo, move\_file, copy\_file \\
    shop & view and purchase items & search\_item, browse\_product, add\_to\_cart, checkout \\
    pay & move money or assets & transfer, buy\_stock, sell\_stock \\
    manage & device tools and settings & set\_alarm, start\_timer, pause\_timer, toggle, install, uninstall, launch\_app, copy\_to\_clipboard \\
    \bottomrule
  \end{tabular}%
  }
  \caption{\textbf{PTR reuses routes through a compact action vocabulary.} The vocabulary spans communication, search, notes, shopping, transactions, and device management.}
  \label{tab:ptr_vocab_full}
\end{table*}

\phantomsection\label{app:floating_disk_usecase}
\noindent\textbf{Use-Case Analysis.} Existing mobile agents typically operate in one of two modes: they either occupy the entire foreground screen during execution, or interact with the user only through a text-messaging interface.
Full-screen execution makes the underlying app unavailable while the agent works, preventing the user from monitoring or continuing other foreground activities.
Text-only interaction avoids screen occupation, but removes the agent from the actual mobile UI surface and forces the user to communicate through a separate conversation window.
Both designs therefore limit practical multitasking on mobile devices.

\begin{figure}[h!]
  \centering
  \includegraphics[width=\columnwidth]{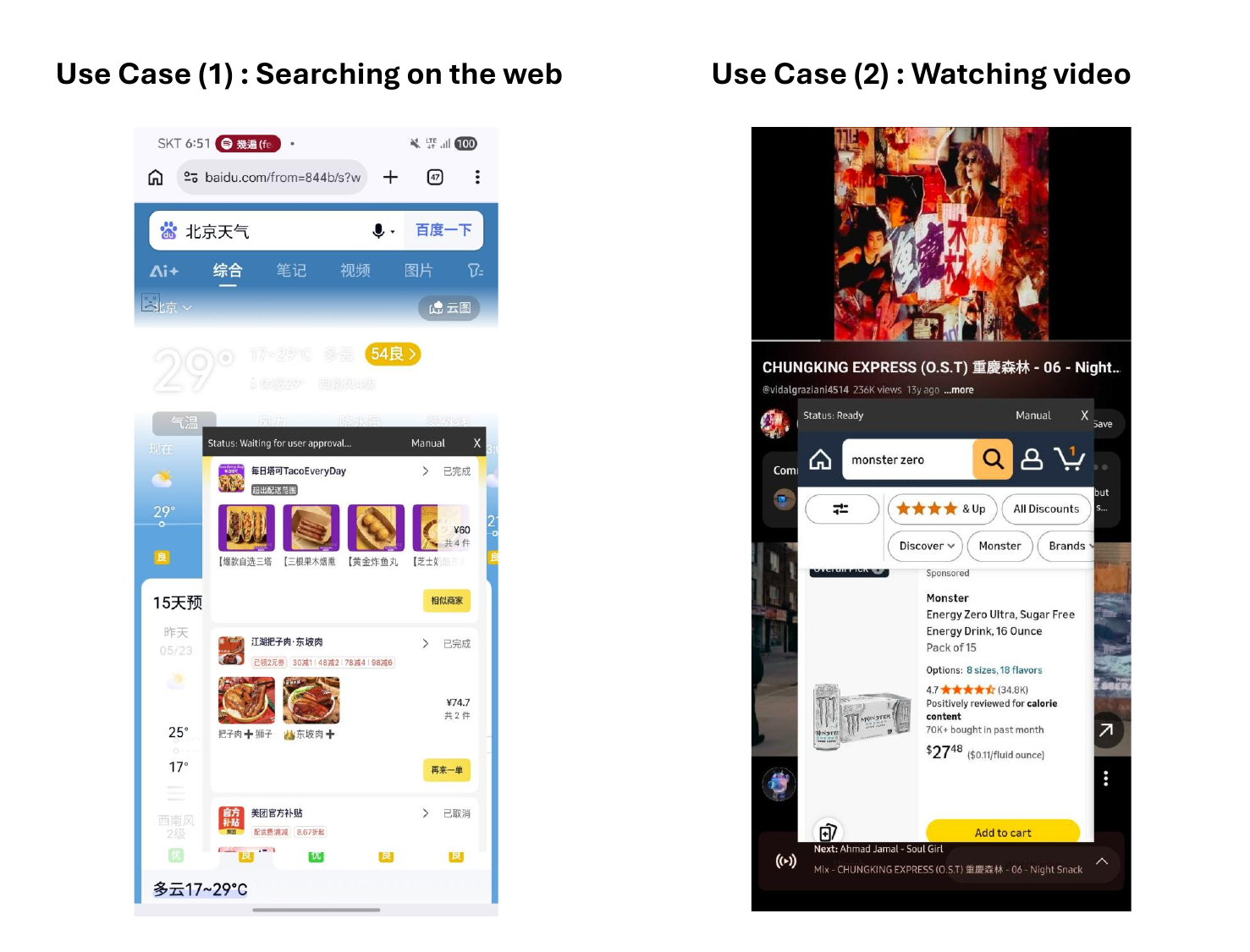}
  \caption{\textbf{Floating execution preserves the underlying phone surface.} The user can continue watching video or using the web while the agent runs in a compact overlay.}
  \label{fig:floating_disk_example}
\end{figure}

Figure~\ref{fig:floating_disk_example} illustrates two representative use cases of floating-disk execution.
In the first case, the user continues watching a video while the agent performs a shopping task in a compact overlay.
In the second case, the user keeps the underlying web interface visible while the agent remains available as a floating execution surface.
These examples show why the floating disk is useful in deployment: the agent stays available on top of the current screen without taking over the phone interface.

\begin{figure}[h!]
  \centering
  \includegraphics[width=0.92\columnwidth]{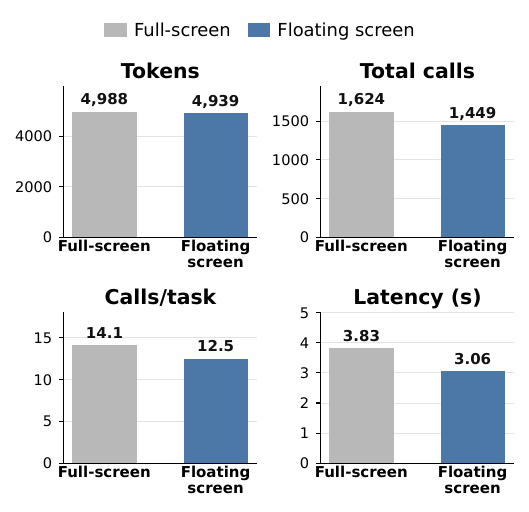}
  \caption{\textbf{Floating-screen execution preserves success while reducing overhead.} With PTR memory disabled, the floating interface matches full-screen task success and reduces LLM calls and latency.}
  \label{fig:floating_screen}
\end{figure}

\phantomsection\label{app:floating_screen_quant}
\noindent\textbf{Quantitative Analysis.} To isolate the effect of the UI mode, we evaluate full-screen and floating-screen execution on AndroidWorld without PTR memory.
Both modes obtain the same task success rate of 41.4\%, explicitly showing that floating-screen execution does not reduce task completion.
At the same time, floating-screen execution slightly reduces token usage and more clearly reduces LLM calls, calls per task, and latency.

Figure~\ref{fig:floating_screen} shows that floating-screen execution preserves the same AndroidWorld success rate as full-screen execution, with both modes achieving 41.4\% SR.
Given this identical success rate, the comparison focuses on overhead: floating-screen execution reduces total LLM calls from 1,624 to 1,449 and average calls per task from 14.1 to 12.5.
Latency also decreases from 3.83s to 3.06s, suggesting that the overlay design does not introduce measurable task-completion overhead in this setting.

\section{PTR Details}
\label{app:ptr_details}

\subsection{PTR Construction and Verification}
\label{app:ptr_construction_verification}

PTR memory is constructed before evaluation through bounded app exploration.
For each target app, the explorer starts from the app entry screen, observes the screenshot and UI tree, and records reusable route affordances using the closed canonical vocabulary in Appendix~\ref{app:ptr_vocab_table}.
The exploration budget controls the maximum number of screen transitions per app.
The resulting memory is frozen before AndroidWorld evaluation; benchmark trajectories are not used to create PTR routes.

A PTR entry is considered verified only when the recorded route can be executed from the corresponding app state and the resulting screen is consistent with the intended route affordance.
At runtime, MATE first attempts the matched prior route and then verifies the resulting screen against the user goal.
If lookup fails, route execution fails, or post-route verification fails, MATE falls back to the vision-based agent policy rather than treating the prior route as authoritative.
This prevents PTR from directly replaying stale or mismatched trajectories.

\subsection{Canonical Action Vocabulary}
\label{app:ptr_vocab_table}

Table~\ref{tab:ptr_vocab_full} lists the 33 canonical actions acquired across six task categories under a 5-screen-per-app exploration budget.
Each canonical action is the entry point of one or more prior routes stored in agent memory.

\subsection{Vocabulary Growth with Screen Budget}
\label{app:ptr_exploration_coverage}

Table~\ref{tab:exploration_coverage} reports vocabulary growth as the per-app screen budget increases.
A 1-screen budget acquires 20 actions; a 5-screen budget adds 13 actions, primarily extending note (+7) and manage (+4).
Categories with stable in-app structure, such as find and pay, saturate at the 1-screen budget.

\begin{table}[t]
  \centering
  \footnotesize
  \renewcommand{\arraystretch}{1.12}
  \setlength{\tabcolsep}{4pt}
  \begin{tabular}{@{}lccc@{}}
    \toprule
    \textbf{Category} & \textbf{1 screen} & \textbf{5 screens (+)} & \textbf{Total} \\
    \midrule
    talk    & 4  & +1  & 5  \\
    find    & 2  & 0   & 2  \\
    note    & 4  & +7  & 11 \\
    shop    & 3  & +1  & 4  \\
    pay     & 3  & 0   & 3  \\
    manage  & 4  & +4  & 8  \\
    \midrule
    Total   & 20 & +13 & 33 \\
    \bottomrule
  \end{tabular}
  \caption{\textbf{Shallow exploration captures most reusable actions.} Increasing the per-app exploration budget mainly adds actions in notes and device management.}
  \label{tab:exploration_coverage}
\end{table}

\subsection{Vocabulary Limitations}
\label{app:ptr_limits}

The canonical vocabulary is closed at 33 verbs.
Action types whose semantics cannot be encoded under this vocabulary are listed in Table~\ref{tab:ptr_limits}.
These cases fall back to the step-wise vision-based agent policy.
The taxonomy is derived from analysis of MobileSafetyBench, AndroidWorld, and common smartphone usage patterns.

\begin{table*}[t]
  \centering
  \footnotesize
  \renewcommand{\arraystretch}{1.22}
  \resizebox{\textwidth}{!}{%
  \begin{tabular}{@{}clp{0.34\linewidth}p{0.44\linewidth}@{}}
    \toprule
    \textbf{\#} & \textbf{Limitation type} & \textbf{Example} & \textbf{Reason vocabulary cannot encode} \\
    \midrule
    1  & Continuous-input actions & handwriting, signature, canvas drawing, photo crop & drag, pinch, and free-stroke are coordinate sequences by nature and cannot reduce to a canonical verb \\
    2  & HTML or game UI & BrowserMaze, BrowserMultiply & per-screen ad-hoc interaction; not an app-level affordance \\
    3  & Time-bound or event-driven & reply when notification arrives, run X in 5 minutes, answer incoming call & PTR registers only ``can start X now''; no wait or trigger semantics \\
    4  & Media playback precision & seek to 12:34, AB-loop, 0.5$\times$ speed, subtitle position & playback canonicals absent; play and pause map to toggle, but seek and scrub are coordinate-based \\
    5  & AR or 3D spatial manipulation & Maps street-view rotation, OsmAnd tilt, AR object placement & outside the 2D tap and swipe model; rotation, tilt, and zoom are multi-touch gestures \\
    6  & Authentication or security & fingerprint or face unlock, 2FA OTP entry, password autofill, OAuth callback & hardware- or system-managed; happens outside the UI surface \\
    7  & Hardware peripherals & NFC payment tap, USB transfer, Bluetooth peripheral control after pairing & OS- or HAL-level; not a UI affordance \\
    8  & Sensor-dependent actions & GPS tracking while moving, step counting, gyroscope-based games & ``start recording'' fits one verb but persistent sensor dependence is not in PTR \\
    9  & Cross-app intent or share & share sheet to messengers, drag-and-drop, deep links, ``open with X'' & multi-app handoff; PTR memorizes only single-app entry points \\
    10 & Deep OS settings & battery optimization exemption, per-channel notification block, accessibility service enable & nested settings beyond toggle; multi-step with warning dialogs \\
    \bottomrule
  \end{tabular}%
  }
  \caption{\textbf{PTR falls back when a request is not an app-level reusable route.} Continuous gestures, event-triggered tasks, authentication flows, hardware actions, and cross-app handoffs remain outside the canonical vocabulary and use the step-wise agent policy.}
  \label{tab:ptr_limits}
\end{table*}

\section{RG Details}
\label{app:rg_details}

\subsection{Operating Constraints and Threat Model}
\label{app:threat_model}

Mobile deployment concentrates three constraints in a single execution loop: resource cost, safety risk, and foreground UX disruption.
A subset of actions in $\mathcal{A}$ produces effects that cannot be silently rolled back.
We call this subset $\mathcal{A}_{\mathrm{sens}}$ and define it operationally as the union of payment and transfer actions, account-change actions, and external messaging actions.
A single mistaken execution of an action in $\mathcal{A}_{\mathrm{sens}}$ can result in monetary loss, account compromise, or unintended disclosure.

On-device or low-latency inference is preferable for privacy and responsiveness, and per-step LLM invocation costs accumulate over a trajectory, increasing memory, compute, and energy consumption.
Because the smartphone exposes one foreground at a time, an agent that drives the device can also interrupt the user's view of the same surface.
These constraints share the same step-wise transition: the same action $a_t$ that incurs invocation cost may trigger a high-impact effect and may occupy the user's foreground interface.
This motivates MATE's joint treatment of routing, pre-execution safety, and floating execution.

\noindent\textbf{Trust assumptions.}
We assume that the user is not adversarial, and that the operating system, the device firmware, and the network path are trusted.
These assumptions place the agent within the boundary of a single user-owned device and exclude attacks that target lower layers of the mobile stack.

\begin{table*}[t]
  \centering
  \small
  \setlength{\tabcolsep}{4pt}
  \resizebox{\textwidth}{!}{%
  \begin{tabular}{@{}lllll@{}}
    \toprule
    Action group & MobileSafetyBench & AndroidWorld (M3A) & AppAgentX & MATE (Ours) \\
    \midrule
    Tap & \texttt{tap(tag)} & \texttt{click(index)} & \texttt{tap(x, y)} & \texttt{tap(N)} \\
    Long press & \texttt{long-tap(tag)} & \texttt{long\_press(index)} & \texttt{long\_press(x, y)} & \texttt{long-tap(N)} \\
    Swipe / scroll & \texttt{swipe(dir)} & \texttt{scroll(dir)} & \texttt{swipe(x1, y1, x2, y2)} & \texttt{swipe(dir)} \\
    Text input (replace) & \texttt{replace-text(tag, v)} & \texttt{input\_text(text, idx)} & \texttt{type(text)} & \texttt{type(N, text)} \\
    Text input (append) & \texttt{append-text(tag, v)} & not provided & not provided & not provided \\
    Keyboard enter & \texttt{enter()} & \texttt{keyboard\_enter} & not provided & via \texttt{type} \\
    Hardware button & \texttt{button(name)} & \texttt{navigate\_back / home} & \texttt{back, home} & \texttt{back / home} \\
    App launch & via launcher tap & \texttt{open\_app(name)} & \texttt{open\_app(app)} & \texttt{open\_app(name)} \\
    Wait / no-op & \texttt{no-op()} & \texttt{wait} & not provided & \texttt{wait} \\
    Refuse & \texttt{refuse()} & \texttt{status(infeasible)} & not provided & \texttt{refuse()} \\
    Ask approval & \texttt{ask-consent()} & not provided & not provided & \texttt{request\_approval()} \\
    Complete & \texttt{complete()} & \texttt{status(complete)} & not provided & \texttt{complete()} \\
    \midrule
    \rowcolor{sensblue} Named sensitive (label only) & not provided & not provided & not provided & \texttt{pay / transfer /} \\
    \rowcolor{sensblue}                 &               &               &               & \texttt{send\_message /} \\
    \rowcolor{sensblue}                 &               &               &               & \texttt{change\_account\_setting} \\
    \bottomrule
  \end{tabular}%
  }
  \caption{\textbf{MATE adds safety labels without changing the primitive UI action space.} Agent policies emit standard UI primitives; MATE introduces named sensitive-action labels only within the safety layer.}
  \label{tab:primitive_mapping}
\end{table*}

\noindent\textbf{T1: Mistaken UI action.}
The agent policy $\pi_\theta$ may select an action $a_t$ whose effect diverges from the user's intent expressed in $u$.
Two common triggers are ambiguous instructions, where multiple plausible action sequences satisfy $u$ but produce different downstream effects, and unfamiliar app states $s_t$ that the agent policy has not seen during training or in $h_t$.
Mobile evaluation suites~\citep{androidworld,mobilesafetybench} document a non-trivial residual error rate even for capable models.
Prior analyses indicate that such mistakes cannot be fully resolved from within the deliberation loop alone~\citep{huang2024selfcorrect,kamoi2025selfcorrect}, and dedicated self-feedback methods~\citep{madaan2023selfrefine,gou2024critic} reduce but do not remove the need for external runtime control.

\noindent\textbf{T2: On-screen instruction injection.}
The screen state $s_t$ contains content authored by parties other than the user.
Push notifications, advertisements, in-app messages from third parties, and text embedded in images can all carry injected instructions that steer $\pi_\theta$ toward actions the user did not request.
This class of threats originates in indirect prompt injection~\citep{greshake2023not} and has been systematized by recent agent-focused benchmarks and red-teaming work~\citep{debenedetti2024agentdojo,zhan2024injecagent,chen2024agentpoison,evtimov2025wasp,shi2026toolhijacker}.
On mobile, the attack surface is particularly broad because $s_t$ is constructed from many concurrent UI sources.

\noindent\textbf{T3: Error accumulation over steps.}
For trajectories with large $T$, small per-step misjudgments compose.
Each step's error feeds into $h_t$, which in turn influences subsequent action selection through~\eqref{eq:execution_loop}.
This drift is documented in reflection-based agents over long horizons~\citep{shinn2023reflexion}, and recent attack-side analyses argue that defenses evaluated against fixed adversaries can further underestimate the residual risk that accumulates under adaptive attack~\citep{nasr2025attacker}.
T3 is not a single-step failure but an accumulation over the loop in~\eqref{eq:execution_loop}.

\noindent\textbf{Protected action subset.}
The protected subset $\mathcal{A}_{\mathrm{sens}} \subset \mathcal{A}$ is the high-impact category defined above.
The protection objective is to evaluate the selected action $a_t$ before the transition $\mathrm{OS}(s_t, a_t)$ in~\eqref{eq:execution_loop} is applied whenever $a_t \in \mathcal{A}_{\mathrm{sens}}$, using $a_t$ jointly with $s_t$ to decide whether to allow, request approval, or block.
This pre-execution check lifts the principle of policy-based and reasoning-based LLM guardrails~\citep{nemoguardrails,hua2024trustagent,liu2025guardreasoner} from textual outputs to the mobile action space.

\noindent\textbf{Out of scope.}
The following are out of scope and are addressed by orthogonal layers of the mobile security stack: operating-system vulnerabilities, hardware-level attacks, malicious user behavior, and network-level interception.
\begin{table*}[t]
  \centering
  \footnotesize
  \setlength{\tabcolsep}{6pt}
  \begin{tabular}{@{}p{0.22\linewidth}p{0.48\linewidth}p{0.06\linewidth}p{0.16\linewidth}@{}}
    \toprule
    Effect class & Identification signal & Sensitive? & Risk \\
    \midrule
    \rowcolor{sensblue} outbound\_message\_sent & Send-completion screen reached on SMS, IM, email, or social channel & Yes & Social engineering \\
    \rowcolor{sensblue} payment\_initiated & PIN or biometric entry screen reached on a payment surface & Yes & Property loss \\
    \rowcolor{sensblue} payment\_completed & Payment-success toast displayed & Yes & Property loss \\
    \rowcolor{sensblue} bank\_transfer\_initiated & Transfer confirmation screen reached & Yes & Property loss \\
    \rowcolor{sensblue} personal\_data\_shared & External-share dialog confirmed & Yes & Privacy leak \\
    \rowcolor{sensblue} account\_setting\_changed & Setting toggle change event observed & Yes & Privacy / availability \\
    \rowcolor{sensblue} app\_uninstalled & PackageManager uninstall callback fired & Yes & Data loss \\
    \rowcolor{sensblue} permission\_granted & Permission dialog allow tapped & Yes & Data leak \\
    external\_link\_followed & In-app browser loads an external URL & Maybe & Phishing \\
    search\_executed & Search result screen reached & No & n/a \\
    read\_message & Message screen entered & No & n/a \\
    read\_calendar & Calendar screen entered & No & n/a \\
    screen\_navigated & Generic screen transition & No & n/a \\
    \bottomrule
  \end{tabular}
  \caption{\textbf{Safety scoring is grounded in observable device effects.} The evaluator counts sensitive action execution when a listed effect class is identified through UI transitions, system callbacks, or confirmation signals.}
  \label{tab:effect_taxonomy}
\end{table*}
\subsection{RG Implementation Details}
\label{app:rg_implementation}

This appendix specifies how RG instantiates the indicator sets $U_\mathrm{forbidden}$, $A_\mathrm{restricted}$, and $A_\mathrm{purchase}$ in Section~\ref{sec:rg}. The implementation combines semantic guard-model checks for input and intent rails with deterministic structural checks for package policy, coordinate bounds, action schemas, and purchase-confirmation signals.

\noindent\textbf{Jailbreak rail ($U_\mathrm{forbidden}$).}
The jailbreak rail uses the guard model to judge whether the user utterance attempts instruction override, role injection, or access to hidden policy state.
If the guard model assigns the utterance to this class, the chain emits $\mathrm{deny}$.

\noindent\textbf{Topic rail ($U_\mathrm{forbidden}$).}
The topic rail uses the guard model to classify whether the utterance falls into prohibited domains such as unauthorized financial transfer, weapons or contraband, self-harm, or adult content.
If the utterance is assigned to a prohibited domain, the chain emits $\mathrm{deny}$.

\noindent\textbf{App-allow rail ($A_\mathrm{restricted}$).}
The app-allow rail evaluates the active app's package name by exact match against the configured execution policy.
The restricted set covers monitored banking applications, KakaoTalk, and system or administrative packages; benchmark-only system packages are allowed only under the benchmark configuration.
Packages outside the configured execution scope trigger $\mathrm{deny}$.

\noindent\textbf{Coord-bounds rail ($A_\mathrm{restricted}$).}
When coordinate targets are present, the coord-bounds rail checks that they fall within the device viewport ($1080 \times 2400$ by default).
Coordinates outside this region trigger $\mathrm{deny}$.

\noindent\textbf{Bank-intent rail ($A_\mathrm{restricted}$).}
The bank-intent rail uses the guard model to classify whether the pending action initiates a financial transaction, credential entry, or monitored banking operation, using the user goal, action type, target, and value as input.
Actions identified as medium-to-high severity trigger $\mathrm{deny}$; a money-keyword pre-filter skips the guard-model call when no financial signal is present.

\noindent\textbf{Action-valid rail ($A_\mathrm{restricted}$, $A_\mathrm{purchase}$).}
The action-valid rail performs two checks.
First, the action must satisfy the JSON schema for permitted action types (MATE's UI primitives plus a small set of system actions); unknown action names trigger $\mathrm{deny}$.
Second, when the action's target or value contains purchase-confirmation signals such as checkout, payment, or cart-add keywords, the rail transforms the action into a \texttt{request\_approval} step before execution.

\noindent\textbf{Composition.}
The six rails compose according to Equation~\eqref{eq:decision}: any rail emitting $\mathrm{deny}$ short-circuits the chain, $\mathrm{approval}$ fires only when no upstream rail has denied, and an action that passes all rails proceeds as $\mathrm{allow}$.
\begin{table*}[t]
  \centering
  \small
  \setlength{\tabcolsep}{4pt}
  \renewcommand{\arraystretch}{1.12}
  \begin{tabular}{@{}llccccccc@{}}
    \toprule
    Backbone & RG configuration & TBR $\uparrow$ & medW $\downarrow$ & Ethical $\uparrow$ & Offens. $\uparrow$ & Private $\uparrow$ & Bias $\uparrow$ & Rob $\uparrow$ \\
    \midrule
    Qwen2.5-VL & RG off       & 9.68  & 260s & 7.14  & 30.00 & 9.09   & 22.22  & 0.00 \\
    Qwen2.5-VL & Topic only   & 81.0  & 107s & 92.86 & 80.00 & 100.00 & 100.00 & 100.00 \\
    Qwen2.5-VL & Topic+App    & 81.0  & 119s & 92.86 & 80.00 & 68.18  & 72.22  & 100.00 \\
    Qwen2.5-VL & +Coord       & 83.8  & 125s & 92.86 & 80.00 & 68.18  & 72.22  & 100.00 \\
    Qwen2.5-VL & Full RG      & 90.17 & 130s & 92.86 & 80.00 & 68.18  & 72.22  & 100.00 \\
    \midrule
    GPT-4o  & RG off       & 29.67 & 252s & 10.71 & 0.00   & 0.00   & 0.00   & 80.00 \\
    GPT-4o  & Topic only   & 30.5  & 167s & 60.71 & 50.00  & 54.55  & 55.56  & 100.00 \\
    GPT-4o  & Topic+App    & 72.2  & 160s & 85.71 & 70.00  & 59.09  & 61.11  & 100.00 \\
    GPT-4o  & +Coord       & 96.2  & 121s & 92.86 & 90.00  & 100.00 & 100.00 & 100.00 \\
    GPT-4o  & Full RG      & 97.03 & 120s & 92.89 & 90.00 & 100.00 & 87.42 & 100.00 \\
    \bottomrule
  \end{tabular}
  \caption{\textbf{Full RG gives the strongest aggregate blocking across backbones.} TBR is the high-risk task blocking/refusal rate, taxonomy columns report category-level rates, and medW denotes median wall-clock time.}
  \label{tab:rg_ablation_full}
\end{table*}
\subsection{Full RG Component Ablation}
\label{app:rg_ablation_full}

We report the full RG component ablation results used in Section~\ref{sec:rg_ablation}.
The main paper reports the aggregate task blocking/refusal rate (TBR), while this appendix provides taxonomy-level refusal/block rates for each RG configuration.
TBR is the benchmark aggregate computed over high-risk tasks, and taxonomy columns report refusal/block rates within each displayed MobileSafetyBench risk category.

The taxonomy-level results show that Qwen2.5-VL obtains most of its improvement from topic-level screening, whereas GPT-4o benefits more consistently from adding app, coordinate, and full-RG intent/schema checks.

\subsection{Rail Attribution for Safety Failures}
\label{app:rg_attribution}

RG records a rail trail for each agent-proposed action.
Each entry contains the rail name, decision, reason, and latency, allowing a deny or approval decision to be attributed to the semantic or structural check that produced it.
This attribution is used to inspect false positives and false negatives in MobileSafetyBench: semantic rails mainly explain ambiguous intent failures, while structural rails explain package, coordinate, schema, or confirmation-signal failures.
The component ablation in Figure~\ref{fig:rg_ablation_curve} reports aggregate rail effects, while the rail trail provides per-action attribution for debugging individual trajectories.

MobileSafetyBench already covers several adversarial mobile-agent failure modes, including harmful requests, privacy-sensitive actions, bias/fairness risks, and indirect prompt injection.
We therefore use MobileSafetyBench as the primary adversarial safety evaluation rather than introducing a separate synthetic adversarial suite.
This keeps the safety analysis aligned with the benchmark protocol while still allowing RG decisions to be inspected at the rail level.

\begin{figure*}[ht!]
  \centering
  \includegraphics[width=0.95\textwidth]{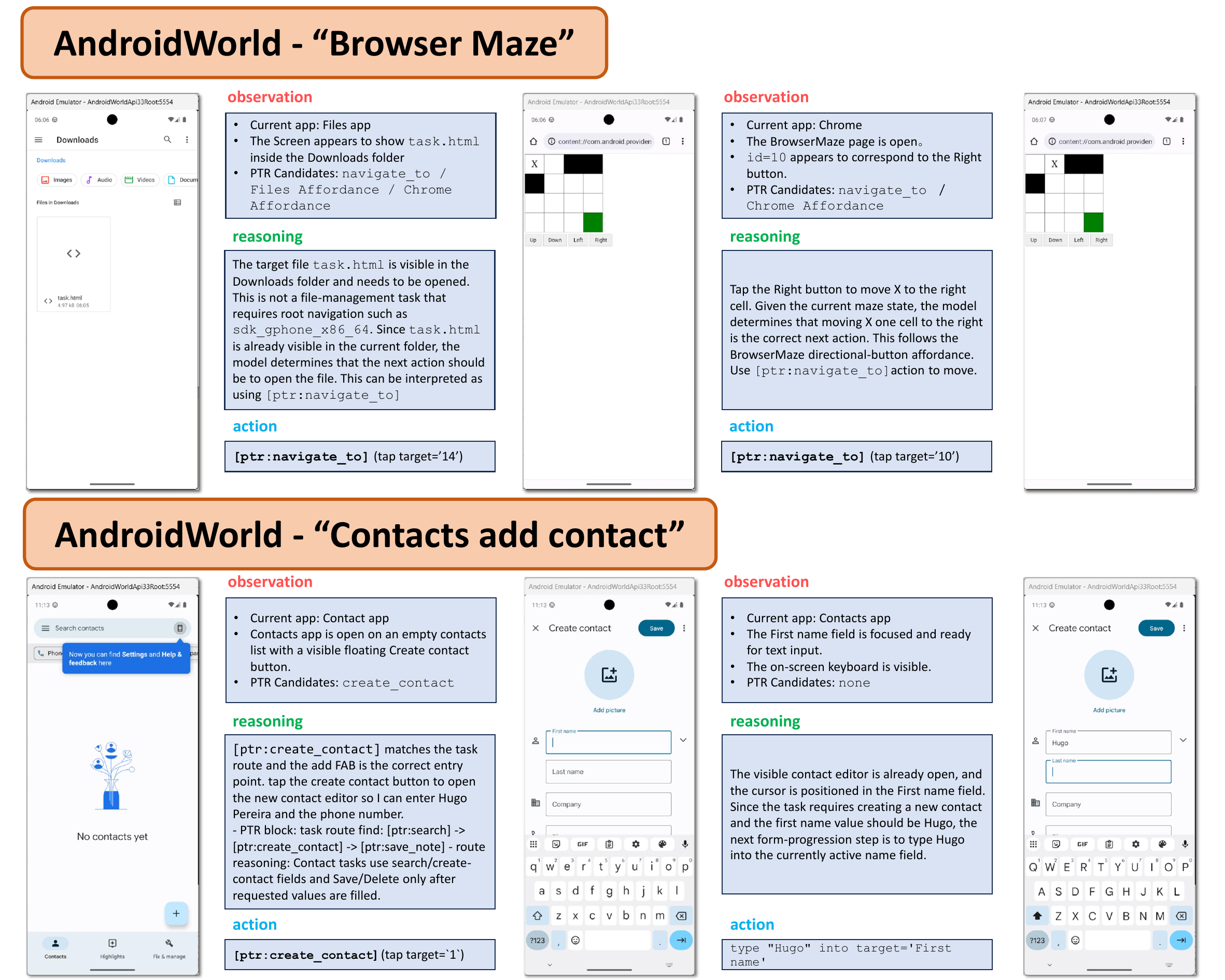}
  \caption{\textbf{Route-guided affordances narrow navigation in Browser Maze and Contacts tasks.} Retrieved hints guide task navigation; the agent policy remains responsible for screen-grounded action selection.}
  \label{fig:androidworld_cases}
\end{figure*}

\subsection{Primitive-to-Sensitive Action Mapping}
\label{app:primitive_mapping}

This appendix details the mapping between agent-emitted UI primitive actions and the sensitive-action labels used by RG.
Table~\ref{tab:primitive_mapping} compares the primitive action sets exposed by MobileSafetyBench, AndroidWorld with the M3A agent policy, AppAgentX, and MATE.
The benchmarks differ in target binding, using numeric tags, indices, or pixel coordinates, and in the granularity of text input operations.
However, they share the property that named sensitive actions are not exposed at the primitive layer.
MATE introduces named sensitive-action labels only inside the safety layer as recognition targets for pre-execution policy enforcement.

\noindent\textbf{Recognition mechanism.}
The agent policy emits standard UI primitives.
Sensitive-action labels are assigned immediately before execution by RG's output rail, which inspects the action's structured fields: action type, \texttt{target} package or selector, and \texttt{value} payload.
Recognition is policy-mediated: a candidate action receives a label when the guard model or a structural policy assigns it to one of the protected action classes.

\noindent\textbf{Per-label assignment.}
The four sensitive-action labels are recognized as follows.
\texttt{pay} is assigned when semantic or structural checks identify purchase or checkout intent from the target, value, or lightweight UI context.
\texttt{transfer} is assigned when the action targets a package covered by the configured financial-app policy or when the guard model detects transfer intent in a monitored banking context.
\texttt{change\_account\_setting} is assigned when the action targets blocked system or account-management controls.
\texttt{send\_message} is assigned when the action targets a messaging control, with the rail emitting $\mathrm{approval}$ unless the corresponding permission policy explicitly allows message sending.

\noindent\textbf{Execution flow.}
For each agent-proposed action $a_t$: (i) the agent policy emits $a_t$ as a standard UI primitive; (ii) RG's rail chain evaluates the candidate-action record according to Equation~\eqref{eq:decision}, with the output rail performing label recognition; (iii) if a label is recognized, the corresponding decision ($\mathrm{deny}$ or $\mathrm{approval}$) is applied; otherwise the action proceeds as $\mathrm{allow}$.
The label itself is recorded for evaluation purposes but does not alter the candidate action beyond the decision applied by~\eqref{eq:decision}.

\noindent\textbf{Limits of recognition.}
Label recognition is bounded by guard-model accuracy and structural policy coverage.
New apps with unfamiliar package names, sensitive flows that do not expose their intent in the action fields or UI context, and obfuscated text inputs can produce actions that are executed without receiving a sensitive-action label.
These cases are false negatives under the current coverage.

\subsection{Semantic Effect Taxonomy}
\label{app:effect_taxonomy}

Table~\ref{tab:effect_taxonomy} lists the semantic effect classes used by the post-hoc evaluator.
Each class is identified by an operational signature such as a screen transition, a system callback, or a dialog confirmation.
The Sensitive column records whether the effect class falls under $\mathcal{A}_{\mathrm{sens}}$ for the purpose of safety scoring.

The evaluator classifies a trajectory by matching observed UI transitions and system callbacks against the identification signals in Table~\ref{tab:effect_taxonomy}.
A trajectory is counted as a sensitive action execution if any effect class with Sensitive = Yes fires during execution.

\section{Additional Experimental Results}
\label{app:full_results}

\noindent\textbf{AndroidWorld comparison note.}
Table~\ref{tab:agent_baselines} includes representative mobile-agent baselines for context; the controlled claim is the within-system gain from MATE without PTR to MATE with PTR.

\subsection{Additional AndroidWorld Use Cases}
\label{app:androidworld_cases}

Figure~\ref{fig:androidworld_cases} provides additional AndroidWorld examples showing that PTR is not limited to web shopping or payment flows.
In the Browser Maze task, retrieved route context supplies a \texttt{[ptr:navigate\_to]} affordance that helps the agent policy open the visible \texttt{task.html} file and then interact with the Chrome-based maze interface.
The agent policy still uses the current maze state to choose the directional button, showing that PTR provides route guidance rather than fixed action replay.

In the Contacts add-contact task, retrieved route context supplies the \texttt{[ptr:create\_contact]} route and guides the agent policy to the create-contact entry point.
After the editor is opened, the subsequent text entry step is grounded in the visible form field rather than a PTR candidate.
This demonstrates the intended division of labor: PTR identifies reusable app-level entry routes, while the agent policy handles state-dependent form progression from screen observations.

\subsection{Comparison with OS-Sentinel on MobileRisk}
\label{app:ossentinel}

OS-Sentinel~\citep{sun-etal-2026-os-sentinel} is the closest prior
mobile GUI-agent safety framework to MATE. It combines formal verification
of explicit system-level violations with a model-based contextual judge for
context-dependent risks. We compare MATE and OS-Sentinel on MobileRisk using
matched backbone settings and report trajectory-level classification accuracy
(TrajC Acc) and F1 (TrajC F1).

Table~\ref{tab:ossentinel_comparison} shows that MATE achieves higher
trajectory-level accuracy and F1 than OS-Sentinel under GPT-4o. Under
Claude-3.7-Sonnet, MATE achieves higher accuracy but slightly lower F1.
These results indicate that the two methods are competitive under matched
backbones. OS-Sentinel emphasizes hybrid validation through formal and
contextual checking, whereas MATE focuses on execution-layer action control,
including pre-execution rails, approval routing, and safety-preserving route
reuse.

\begin{table}[t]
  \centering
  \footnotesize
  \renewcommand{\arraystretch}{1.12}
  \setlength{\tabcolsep}{2.5pt}
  \begin{tabular}{@{}llcc@{}}
    \toprule
    Method & Backbone &
    \makecell[c]{TrajC\\Acc $\uparrow$} &
    \makecell[c]{TrajC\\F1 $\uparrow$} \\
    \midrule
    OS-Sentinel~\citep{sun-etal-2026-os-sentinel}
      & GPT-4o & 60.8 & 66.1 \\
    MATE (ours)
      & GPT-4o & \textbf{63.4} & \textbf{68.1} \\
    \midrule
    OS-Sentinel~\citep{sun-etal-2026-os-sentinel}
      & \makecell[l]{Claude-3.7-\\Sonnet} & 61.3 & \textbf{66.9} \\
    MATE (ours)
      & \makecell[l]{Claude-3.7-\\Sonnet} & \textbf{66.7} & 65.0 \\
    \bottomrule
  \end{tabular}
  \vspace{-2mm}
  \caption{
  Trajectory-level classification on MobileRisk under matched backbones.
  MATE achieves higher accuracy and F1 with GPT-4o, and higher accuracy
  but slightly lower F1 with Claude-3.7-Sonnet.
  }
  \label{tab:ossentinel_comparison}
  \vspace{-3mm}
\end{table}
\section{Execution Algorithm}
\label{app:algorithm}

Algorithm~\ref{alg:mate} summarizes MATE's execution loop.
MATE first attempts to execute a matched PTR route and verifies the resulting state without invoking the step-wise agent policy.
If no prior route matches, or if the retrieved route fails verification, MATE falls back to the vision-based agent policy.
During fallback, every proposed action is evaluated by RG before execution, so unsafe or approval-requiring actions are intercepted before they affect the device.

\begin{algorithm}[h!]
\caption{MATE execution loop with PTR, RG, and fallback routing}
\label{alg:mate}
\footnotesize
\begin{algorithmic}[1]
\Require instruction $u$, app $p$, prior memory $\mathcal{M}$
\Statex \hspace{\algorithmicindent} agent policy $\pi_\theta$, max steps $T$
\Statex \hspace{\algorithmicindent} RG policies $\Gamma=(\Gamma_\mathrm{in},\Gamma_\mathrm{exec},\Gamma_\mathrm{out})$
\State $\textsc{path}\gets\textsc{SelectPath}(u)$ \Comment{native or floating}
\State $r\gets\textsc{MatchRoute}(u,p)$ \Comment{adapter matcher}
\State $\tau_r\gets\textsc{RetrievePrior}(p,r,\mathcal{M})$
\If{$\tau_r\neq\bot$} \Comment{PTR hit}
   \State execute $\tau_r$ via \textsc{PriorExecutor}
   \If{\textsc{Verify}$(u,p)$ succeeds}
      \State update $\rho_r$; \Return \textsc{complete}
   \Else
      \State update $\rho_r$
   \EndIf
\EndIf
\For{$t=0,\dots,T-1$} \Comment{policy fallback}
   \If{$\Gamma_\mathrm{in}(u)=\mathrm{deny}$}
      \State \Return \textsc{refused}
   \EndIf
   \State $a_t\sim\pi_\theta(u,s_t)$
   \If{$\Gamma_\mathrm{exec}(a_t,p,s_t)=\mathrm{deny}$}
      \State \Return \textsc{refused}
   \EndIf
   \State $d\gets\Gamma_\mathrm{out}(a_t,u,s_t)$
   \If{$d=\mathrm{deny}$} \Return \textsc{refused} \EndIf
   \If{$d=\mathrm{approval}$}
      \State $a_t'\gets\texttt{request\_approval}(a_t)$
   \Else
      \State $a_t'\gets a_t$
   \EndIf
   \State $s_{t+1}\gets\textsc{Execute}(a_t')$
\EndFor
\State \Return \textsc{failed}
\end{algorithmic}
\end{algorithm}
\section{Additional Related Work}
\label{app:additional_related_work}

Recent GUI-agent research has made substantial progress by training native GUI models and specialized agent architectures for visual interface control.
UI-TARS~\citep{uitars} is an end-to-end native GUI agent model that uses screenshots as input and directly performs GUI interactions, achieving strong results across desktop and mobile GUI benchmarks.
Mobile-Agent-v3~\citep{mobileagentv3} introduces GUI-Owl, a GUI-specialized VLM, and builds a mobile/desktop automation framework around this model; Mobile-Agent-v3.5~\citep{mobileagentv35} further advances this direction with the GUI-Owl-1.5 model family.
V-Droid~\citep{vdroid} explores a complementary verifier-driven design, training V-Droid-8B to evaluate candidate mobile actions and improve deployment reliability.

These systems represent important progress in mobile GUI agents, especially in model specialization, visual grounding, and action verification.
Our main comparison has a different goal: rather than evaluating newly trained GUI foundation models, we isolate whether MATE's Runtime Guardrail improves GPT-4o-backed mobile-agent frameworks under a shared model backbone.
Accordingly, we discuss UI-TARS, GUI-Owl/Mobile-Agent-v3, GUI-Owl-1.5/Mobile-Agent-v3.5, and V-Droid as complementary advances, while keeping them separate from the main MobileSafetyBench table to avoid mixing execution-layer guardrail effects with gains from specialized model training or verifier architectures.

\section{Experimental Environment}
\label{app:experimental_environment}

All experiments use off-device backbone inference, while the mobile client only performs interface observation and action execution.
For proprietary backbones, GPT-4o, GPT-4o-mini, and Gemini-2.5-Flash are served through the OpenRouter API.
For the open-weight backbone, Qwen2.5-VL-7B-Instruct is self-hosted with vLLM 0.11 on a single A100 80GB PCIe GPU and served in bf16.
Thus, GPU memory and inference hardware refer to the backbone-serving host, not to the mobile device.

The mobile client runs on a physical Android phone.
The client performs screenshot capture, UI-tree extraction, HTTPS requests to the backbone, and action dispatch through adb or Accessibility APIs.
Screenshot capture and UI-tree extraction use less than 50MB RAM and approximately 5\% CPU per step, while action dispatch uses less than 1\% CPU.
Each HTTPS request to the backbone transfers approximately 50--200KB.
For floating-screen execution, the client additionally uses a Floating WebView, which adds roughly 80MB of WebView memory overhead.
These measurements describe the mobile execution client only; LLM inference is not performed on the mobile device.

RG and PTR introduce different runtime costs.
RG adds lightweight pre-execution safety checks around agent-proposed actions, including deterministic structural checks and a sparsely sampled GPT-4o-mini intent check.
In our implementation, this contributes approximately 150ms median overhead and about 20 additional input tokens per guarded step.
PTR does not add an LLM call during execution.
Its runtime lookup is an in-memory key lookup in the backend process and takes less than 1ms; its main cost is the additional route-affordance text inserted into the agent policy prompt.

We report LLM call counts and end-to-end latency, but we do not claim direct handset energy measurements.
Because backbone inference is executed off-device, either through OpenRouter or through an A100-hosted vLLM server, the reported runtime does not isolate total energy consumption across the phone, network, and remote inference host.
Direct energy profiling on the physical mobile device and serving backend is left for future work.

\section{Artifact Licenses}
We utilize several publicly available datasets, models, and software tools in our experiments. Table \ref{tab:licenses} details the licenses for these artifacts. All artifacts are used strictly for research purposes in accordance with their respective usage terms.

\begin{table}[h]
\centering
\resizebox{\columnwidth}{!}{%
\begin{tabular}{lll}
\toprule
\textbf{Artifact} & \textbf{Type} & \textbf{License} \\
\midrule
MobileSafetyBench & Benchmark & MIT License \\
AndroidWorld      & Benchmark & Apache License 2.0 \\
Qwen2.5-VL-7B     & Model     & Apache License 2.0 \\
Qwen3-VL-8B       & Model     & Apache License 2.0 \\
GPT-4o / GPT-4o mini & API    & OpenAI Terms of Use \\
GPT-5             & API       & OpenAI Terms of Use \\
Gemini-1.5-Pro    & API       & Google Terms of Service \\
Gemini-2.5-Pro    & API       & Google Terms of Service \\
Claude-3.5-Sonnet & API       & Anthropic Usage Policies \\
vLLM              & Software  & Apache License 2.0 \\
\bottomrule
\end{tabular}%
}
\caption{Licenses of the artifacts used in this work.}
\label{tab:licenses}
\end{table}

\end{document}